\documentclass[a4paper,conference]{IEEEtran}
\ifCLASSINFOpdf
  \usepackage[pdftex]{graphicx}
\else
\fi
\usepackage{amsmath}
\ifCLASSOPTIONcompsoc
  \usepackage[caption=false,font=normalsize,labelfont=sf,textfont=sf]{subfig}
\else
  \usepackage[caption=false,font=footnotesize]{subfig}
\fi
\usepackage{url}
\begin{document}

\newpage

\title{Dynamic Data Augmentation with Gating Networks for Time Series Recognition}

\author{\IEEEauthorblockN{Daisuke Oba, Shinnosuke Matsuo, and Brian Kenji Iwana}
\IEEEauthorblockA{Department of Advanced Information Technology\\
Kyushu University\\
Fukuoka, Japan\\
Email: iwana@ait.kyushu-u.ac.jp}}

\maketitle

\begin{abstract}
Data augmentation is a technique to improve the generalization ability of machine learning methods by increasing the size of the dataset. However, since every augmentation method is not equally effective for every dataset, you need to select an appropriate method carefully. We propose a neural network that dynamically selects the best combination of data augmentation methods using a Gating Network and a mutually beneficial feature consistency loss. The Gating Network is able to control how much of each data augmentation is used for the representation within the network. The feature consistency loss gives a constraint that augmented features from the same input pattern should be in similar. In the experiments, we demonstrate the effectiveness of the proposed method on the 12 largest time-series datasets from 2018 UCR Time Series Archive and reveal the relationships between the data augmentation methods through analysis of the proposed method.
\end{abstract}


% \fi
\IEEEpeerreviewmaketitle

\section{Introduction}
\label{sec:intro}

% data augmentation 
There have been many successes in time series and signal recognition using neural networks~\cite{Schmidhuber_2015, IsmailFawaz2019DeepLF}. 
Part of the success of neural networks is due to the increase in data~\cite{Banko_2001}. 
However, public time-series datasets are often very small~\cite{UCRArchive2018}. 
One method of overcoming problems with small datasets is the use of data augmentation (DA)~\cite{iwana2021an}. DA is a technique that increases the generalization ability of pattern recognition methods by generating synthetic data to supplement the training set~\cite{Shorten_2019}. 

\begin{figure}%[h]
\begin{center}
\scalebox{0.35}[0.35]{\includegraphics{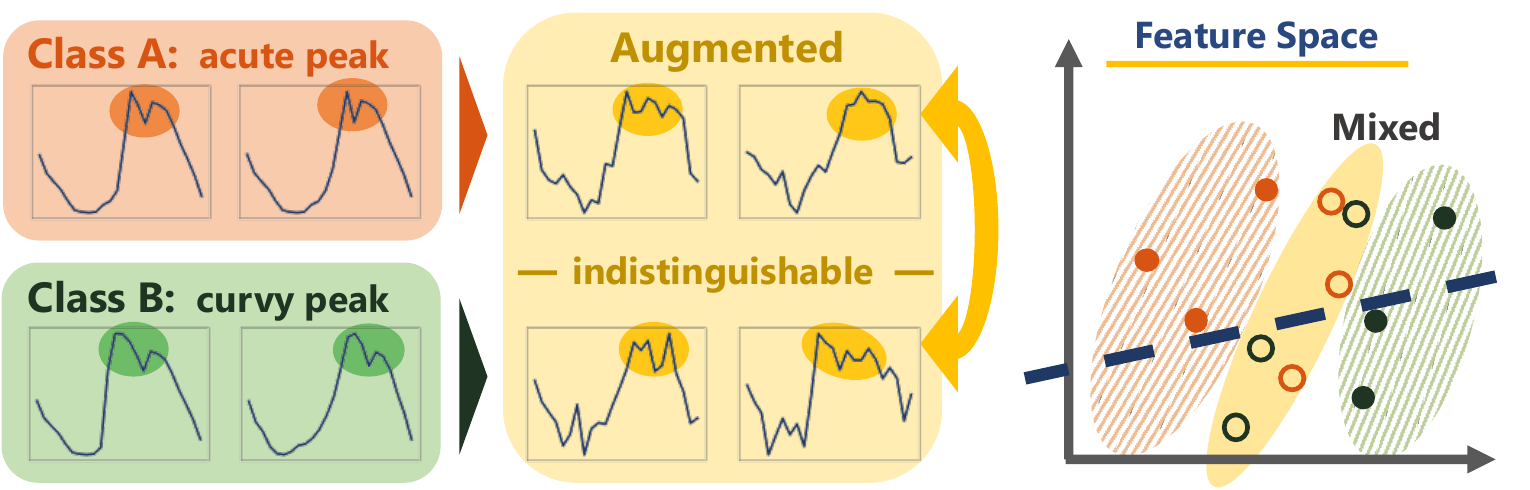}}
\caption{Example of a possible problem using DA with time series. If the DA method causes the time series to lose class-unique features, the distribution of the patterns in the feature space will be mixed. In this example, both the acute peaks in class A and the curvy peaks in class B are lost after Jittering is applied. In the graph on the right, the solid points are original patterns, and the hollow points are augmented patterns. Augmented samples mixed in feature space interfere with the model's ability to draw a boundary.}
\label{fig:problem}
\end{center}
\end{figure}

% difficulty (Fig.1)
Many time series DA methods are random transformations adapted from image recognition. 
However, there is a diverse amount of time series, and each dataset has different properties. 
Thus, not every DA method is suitable for every time series~\cite{iwana2021an}. 
Fig.~\ref{fig:problem} shows an example of how an inappropriate DA method could interfere with recognition.
In the figure, the augmentation obfuscates the discriminating features, making it challenging to define an effective decision boundary. 
% Indeed, DA can give a worse accuracy than without it.
Hence, for each time series dataset, appropriate DA methods must be carefully selected.

% auto augmentation
One solution is the use of automatic DA. 
Automatic DA, such as AutoAugment~\cite{Niu_2019}, attempts to automatically learn the optimal set of DA methods for a model. 
Automatic DA can be generally categorized into three groups: test-time augmentation, training-time augmentation, and methods that use both. 
Test-time augmentation applies DA to the test set and uses an ensemble of predictions to improve generalization. 
Conversely, training time augmentation used augmented training data to train the model. 
Hybrid methods use both test-time and training-time augmentation.

\begin{figure}%[ht]
\begin{center}
\scalebox{0.3}[0.3]{\includegraphics{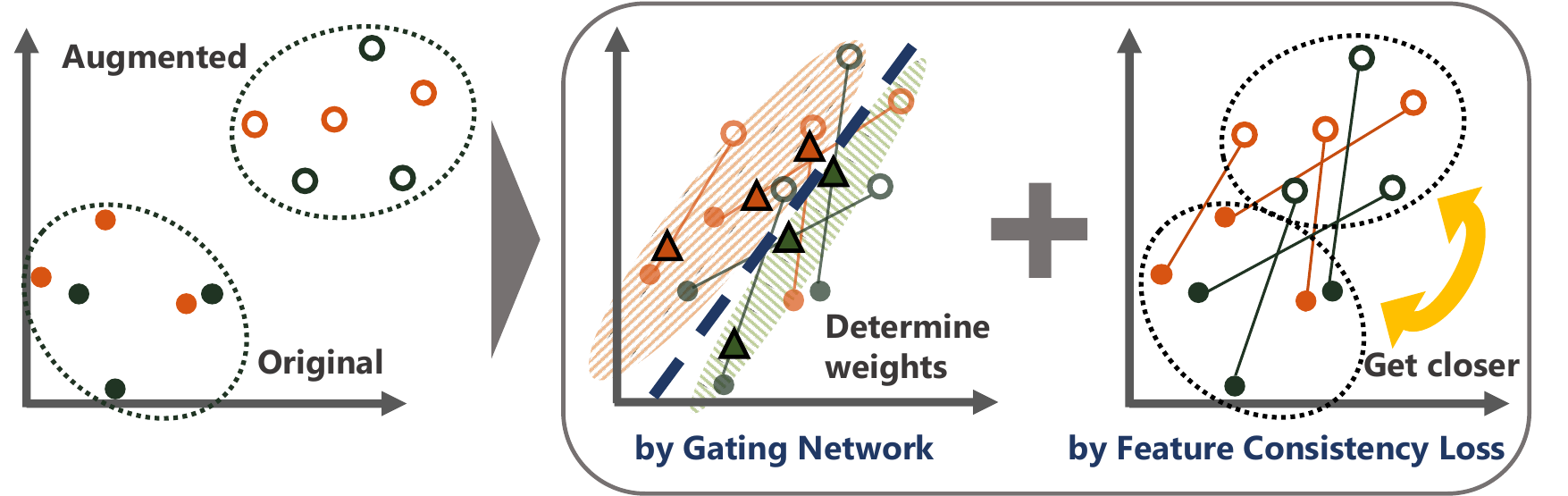}}
\caption{The proposed solution. (left) an example feature space with two classes. The solid points are the original time series, and the hollow points are augmented. (center) how the original and augmented features are combined with the Gating Network. The triangles are the combined features.  (right) the effects of the proposed feature consistency loss.}
\label{fig:solution}
\end{center}
\end{figure}

% proposed method
In this paper, we propose a neural network that selects the best ratio of DA methods.
During training, each DA method is applied to the input data and is fed to the model. 
Inspired by Matsuo et al.~\cite{matsuo2021}, who use a Gating Network to utilize two modalities,
we design a model where each DA method acts as a modality and is combined with the aid of a Gating Network. 
As illustrated in Fig.~\ref{fig:solution}, the features of each stream are weighted by the Gating Network and added. 
In addition, a proposed feature consistency loss is used to give a soft constraint that matches the features of each DA method. This is to ensure the representation of each DA method is similar enough not to cause conflicts.

The contributions of this paper are as follows:
\begin{itemize}
    \item We propose the use of a Gating Network in order to select the best combinations of DA methods automatically. At the same time, we also introduce a feature consistency loss to relate the features of each DA method for the Gating Network.
    \item One benefit of the proposed model is the explainable ability of the learned weighting parameter of the Gating Network. We analyzed this parameter and discovered the meaningful relation between it and the sample distribution.
    \item We demonstrate that the proposed model performs better than most competitors in many cases when tested on 12 datasets from the 2018 UCR Time Series Archive~\cite{UCRArchive2018}.
\end{itemize}

The code is publicly available at https://github.com/uchidalab/DynamicDA-with-GatingNetworks.
%https://github.com/uchidalab/.

\section{Related Work}
\label{sec:related}

\subsection{Data Augmentation for Time Series}
% Data is important, and its availability is increasing
%In machine learning, both the quality and quantity of data play a crucial role for a model to get a generalization ability.
In machine learning, preparing a sufficient amount of data is frequently more effective than elaborating on a model architecture to achieve better performance~\cite{Cubuk_2019}.
Fortunately, data collection and their handling are getting easier and easier nowadays owing to the advances in hardware~\cite{Jouppi2017IndatacenterPA}.

% Data Augmentation in general
However, it is not always assured that you can get a sufficient amount of data to build accurate models.
When encountering a data shortage problem, data augmentation (DA) can be applied~\cite{Tanner1987TheCO}.
Using DA, the dataset size is increased by augmenting original samples with generated samples. 
% It makes additional samples having different features from the originals while conserving the sample's semantic information.

% DA for Time-Series
DA is a technique that can be applied to any type of data, including time series.
Different from other data types, time series have unique characteristics such as trend, seasonality, and outliers.
Effective DA methods transform data preserving class-unique shapes that are often apparent in these characteristics.
There are various DA methods in time series, and most of them can be categorized into two groups: methods distort data in magnitude direction and methods distort data in time direction.

%DA in magnitude
Jittering~\cite{Bishop1995TrainingWN,an1996}, Scaling~\cite{Um_2017}, and Magnitude Warping~\cite{Um_2017} are basics among methods in magnitude direction.
Jittering adds noise to data. This is a simple but still effective to augment data.
Scaling is the method to change the magnitude scale to a certain ratio over the data length.
Magnitude Warping warps data by a smoothed curve.

%DA in time
Methods in the time distortion group displace steps of data to different time steps.
For example, Slicing~\cite{Um_2017} cuts the original data and widens it to the initial steps.
Permutation~\cite{Um_2017} divides data into several segments and rearranges them. One major drawback of this method is the transformed data loses its trend and seasonality easily as its shape is changed without considering the relationship among segments.
Time Warping~\cite{Um_2017,le2016data,rashid2019timewarp} changes data in time direction by a smoothed curve like Magnitude Warping.

\subsection{Automatic Data Augmentation}
Recently, there has been a rise in methods of automatic DA, which makes a model learn how to augment data automatically to get better performance.
% This automation is worthwhile because making a policy on augmenting data requires both professional experiences in machine learning and a deep understanding of target data.
%
Many of these methods are policy-based for image recognition. 
As a definitive example, AutoAugment~\cite{Cubuk_2019} approaches finding the best DA methods by formulating the problem as a discrete search problem. 
It uses reinforcement learning to train a search network to determine the best augmentation policy. 
More recently, there have been improvements on AutoAugment, such as Fast AutoAugment~\cite{lim2019fast} that uses density matching to reduce the search space and Faster AutoAugment~\cite{Hataya_2020} that uses a differential solution to finding the policy. 
RandAugment~\cite{Cubuk_2020} is an even more efficient automatic augmentation method that uses a random policy and a reduced search space.
In addition, Ho et al.~\cite{ho2019population} proposed using a population-based policy. 
There have also been automatic augmentation methods proposed for point clouds~\cite{Li_2020} and modality invariant latent space~\cite{cheung2020modals}.

The previous methods use DA to improve the generalization of the model during training. 
Conversely, test-time DA~\cite{shanmugam2020and} methods use augmented data during the inference step. 
Typically, these methods use multiple DA transformations on the test samples and average the predictions in order to add robustness.

Compared to image recognition, relatively few automatic DA methods are used for time series and sequences. 
In one example, Niu et al.~\cite{Niu_2019} adapted AutoAugment for natural language processing (NLP) by replacing the image augmentation methods with NLP-based methods. 
Sample Adaptive Policy Augmentation (SapAugment)~\cite{Hu_2021} applies DA to each sample weighted by the training loss of the sample. 
Notably, SapAugment was used for speech recognition. 
Fons et al.~\cite{fons2021adaptive} proposed two automatic methods, W-Augmentation and $\alpha$-trimmed Augmentation. 
Both methods apply a variety of time series DA methods to each of the input samples and train the network based on combining the losses. 
% The difference between the two is that W-Augmentation uses a trained weight to weight the losses, and $\alpha$-trimmed Augmentation removes the highest and lowest loss before combination.

The advantage of the proposed method over many of the methods from literature is the ease of the end-to-end training, the need for relatively few hyperparameters, and the explainability. 
Compared to the reinforcement learning-based methods, the proposed method does not require splitting the dataset into subsets and is much easier to build and train. 
And, compared to the loss-based methods, the proposed model has a direct method of observing the usefulness of the DA method through the learned parameter.

\section{Dynamic Data Augmentation}
\label{sec:method}

\begin{figure}[t]
\begin{center}
\scalebox{0.28}[0.26]{\includegraphics{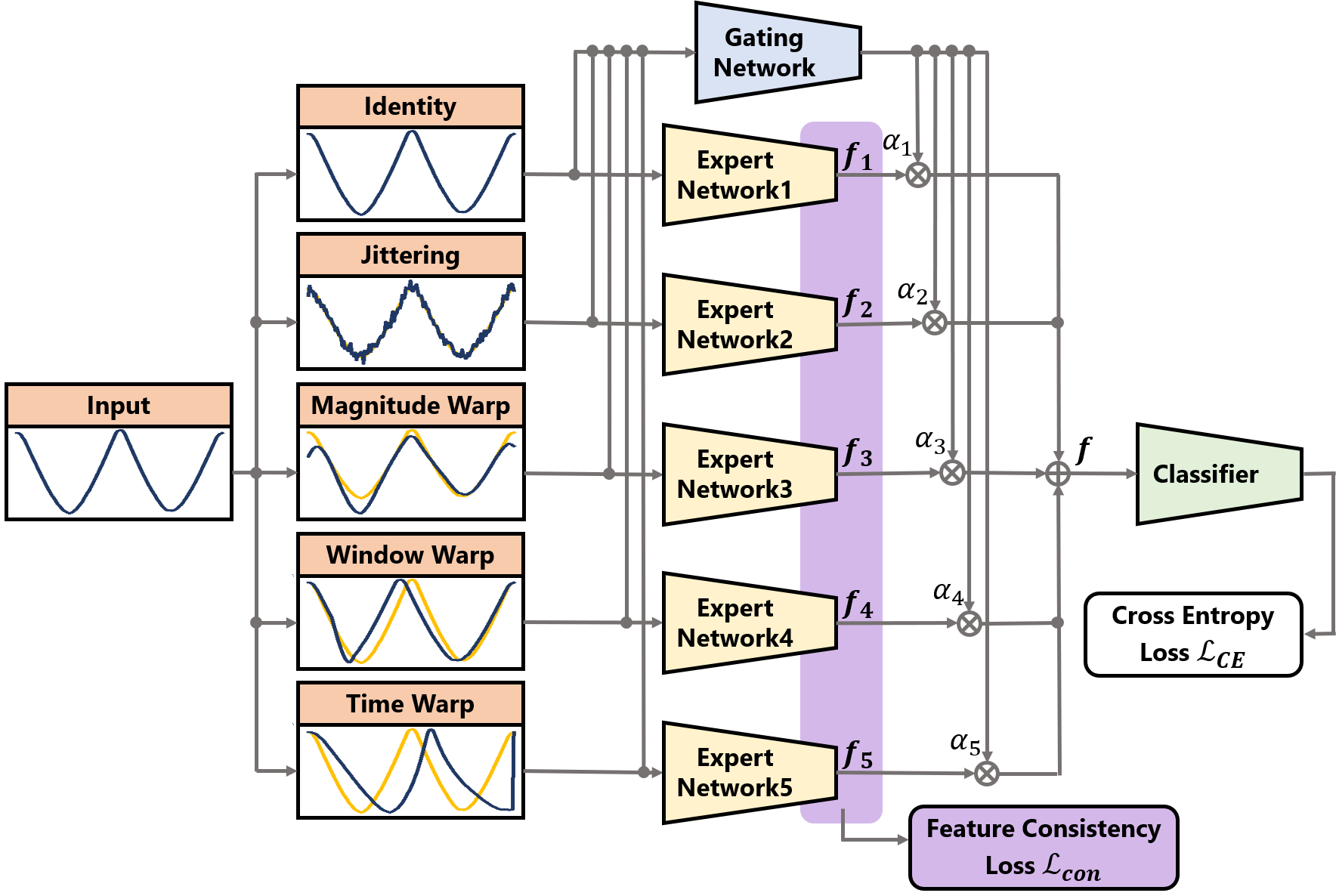}} %0.3&0.28
\caption{An overview of the proposed model $(N=5)$}
\label{fig:overview}
\end{center}
\end{figure}

To realize automatic DA, we designed the model architecture illustrated in Fig.~\ref{fig:overview}.
This network takes a time series input and dynamically adjusts the use of DA methods.
During training time, the model learns how to weigh the DA methods to classify input data class effectively.
With these learned parameters, it gives the best ratios of DA methods for each sample in test time.

During the forward step, the input is transformed by multiple different DA methods.
The transformed data is then sent to two modules: multiple Expert Networks, each corresponding to a different DA method, and a Gating Network to weigh the combination of them.
We also introduce a feature consistency loss to assure each Expert Network's output be in a similar location in feature space to allow for their combination.
Then, the combined features are sent to a Classifier, which returns a prediction.

\subsection{Data Augmentation Methods and Expert Networks}

We employed five basic and comprehensive DA methods.
Jittering and Magnitude Warping are magnitude-based methods, and Time Warping and Window Warping change the time series in the time dimension.
The following is a short description of each of the methods and the hyperparameters used:
\begin{itemize}
    \item \textit{Identity}: The identity Expert Network is the original time series with no augmentation.
    \item \textit{Jittering}: Following Um et al.~\cite{Um_2017}, Jittering adds Gaussian noise with a mean $\mu=0$ and standard deviation $\sigma=0.03$ to the time series.
    \item \textit{Magnitude Warping}: For Magnitude Warping, the time series is multiplied by a smooth curve defined by a cubic spline with knots at random locations and magnitudes. Also following Um et al.~\cite{Um_2017}, four knots are used with $\mu=0$ and $\sigma=0.2$.
    \item \textit{Time Warping}: Time Warping is similar to Magnitude Warping, except the warping is done in the time domain. Also similarly, four knots are used with random magnitudes with $\mu=1$ and $\sigma=0.2$~\cite{Um_2017}.
    \item \textit{Window Warping}: Window Warping selects a random window of 10\% of the original time series length and warps the window by 0.5 or 2 times~\cite{le2016data}. In order to normalize the length for batch training, the warped time series is resampled back to the original length~\cite{iwana2021an}.
\end{itemize}

The transformed data by each method is sent to both the Gating Network and Expert Networks.
The role of Expert Networks is extracting features of transformed data. 
This network consists of a temporal CNN and fully-connected layers.
We attached the fully-connected layers to utilize the global information of data by allowing each node in a fully-connected layer connected with all nodes from the final convolutional layer.

\subsection{Gating Network}

The Gating Network is an additional network used to control each Expert Network's output. 
The input of the Gating Network is the DA transformations concatenated by the channel dimension, and the output is learned weights $\alpha_1,\dots, \alpha_n, \dots \alpha_N$ corresponding to a different DA method, where $N$ is the number of Expert Networks.
Namely, the features $\boldsymbol{f}_n$ of each Expert Network are combined into $\boldsymbol{f}$, or:
\begin{eqnarray}
    \boldsymbol{f} = \sum\limits_{n=1}^N \alpha_n \boldsymbol{f}_n
    \qquad [\alpha_1 + \dots + \alpha_n + \dots + \alpha_N = 1].
\end{eqnarray}
The idea is an extra Gating Network with information from each of the DA methods is used to weigh the importance of each Expert Network for the internal representation for the Classifier.

The weighted features $\boldsymbol{f}_n$ are added and sent to the Classifier and give predictions.
It should be noted that the addition operation is only possible because the whole network is trained end-to-end and with the help of the feature consistency loss (Section~\ref{sec:consitancy}).

\subsection{Feature Consistency Loss}
\label{sec:consitancy}

The features from each Expert Network may be dramatically different across the DA methods. 
It could cause the representation of the DA methods to interfere with each other. 
Therefore, we introduce a feature consistency loss  $\mathcal{L}_\mathrm{con}$.
The feature consistency loss is a soft constraint that encourages the features $\boldsymbol{f}_n$ of each Expert Network to be similar, or:
\begin{eqnarray}
    \mathcal{L}_\mathrm{con} = \frac{1}{N}\sum\limits_{i=1}^N \left\|\boldsymbol{f}_n - \bar{\boldsymbol{f}}\right\|^2,
\end{eqnarray}
where $\boldsymbol{f}_n$ is each node of Expert Network's output and $\bar{\boldsymbol{f}}$ is the mean of them. 
We use this soft constraint instead of a hard constraint because a soft constraint still allows for supplemental information from each Expert Network.
% Therefore, this loss function can be seen as a variance of expert networks' output feature.

During training, the whole network is trained with a combined feature consistency loss $\mathcal{L}_\mathrm{con}$ and cross entropy loss $\mathcal{L}_\mathrm{CE}$ for classification. 
To control the amount of $\mathcal{L}_\mathrm{con}$ used in the total loss $\mathcal{L}_\mathrm{total}$, a hyperparameter $\lambda$ is used, or:
\begin{eqnarray}
    \mathcal{L}_\mathrm{total} = \mathcal{L}_\mathrm{CE} + \lambda\ \mathcal{L}_\mathrm{con}.
\end{eqnarray}
% This is to balance between the feature consistency loss and a cross entropy loss.

\section{Experimental Results}
\label{sec:experiment}

\subsection{Datasets}

\begin{table}%[!t]
    \renewcommand{\arraystretch}{1.3}
    \caption{Dataset Details}
    \label{tab:dataset}
    \centering
    \begin{tabular}{lccccc}
    \hline
    Dataset & Type & Length & \# Class & \# Train & \# Test\\ 
    \hline
    Crop  & Image & 46 & 24 & 7,200 & 16,800 \\ 
    ElectricDevices & Device & 96 & 7 & 8,926 & 7,711 \\ 
    FordA & Sensor & 500 & 2 & 3,601 & 1,320 \\
    FordB & Sensor & 500 & 2 & 3,636 & 810 \\  
    HandOutlines & Image & 2,709 & 2 & 1,000 & 370 \\ 
    MelbournePed & Traffic & 24 & 10 & 1,194 & 2,439 \\ 
    NonInFECGTx1 & ECG & 750 & 42 & 1,800 & 1,965 \\ 
    NonInFECGTx2 & ECG & 750 & 42 & 1,800 & 1,965 \\ 
    PhalangesOC & Image & 80 & 2 & 1,800 & 858 \\ 
    StarLightCurves & Sensor & 1,024 & 3 & 1,000 & 8,236 \\ 
    TwoPatterns & Simulated & 128 & 4 & 1,000 & 4,000 \\ 
    Wafer & Sensor & 152 & 2 & 1,000 & 6,164 \\ 
    \hline
    \end{tabular}
\end{table}

From 2018 UCR Time Series Archive~\cite{UCRArchive2018}, we selected all of the datasets which have $1,000$ or more training samples. 
Through this, 12 datasets are used, and among them, four are sensor data, one is device data, two are electrocardiogram (ECG), three are shape outlines, one is traffic data, and one is simulated data. 
Details of the datasets are shown in Fig.~\ref{tab:dataset}.
We used the pre-defined, provided training and test sets. 
Every dataset was min-max normalized so that the training set is within the range of -1.0 to 1.0. 
For the datasets with variable lengths, zero-padding was used.

\subsection{Implementation Details}

While both Recurrent Neural Network (RNN) based networks and temporal Convolutional Neural Networks (CNN) have been used for time series classification, temporal CNNs have been shown to perform especially well~\cite{Wang_2017,bai2018empirical}.
Thus, we use a temporal CNN for the base of the proposed method.
For the Expert Networks, temporal CNNs with three 1D convolutional layers and two fully-connected layers are used. 
The convolutional layers have 32, 64, and 128 filters, respectively, and the fully-connected layers have 512 nodes each.
Between each convolutional layer, batch norm~\cite{ioffe2015batch}, a rectified linear unit (ReLU) activation, and max pooling is used. 
The Gating Network has an additional fully-connected layer with softmax, where the number of nodes equals the number of Expert Networks $N$. We keep $N=5$ in this work.
The Classifier is a final two fully-connected layers with 512 nodes and the number of classes, respectively.

\subsection{Comparative Evaluations}

In order to evaluate the proposed method, the following four comparative methods are used: one test-time method and three automatic DA methods. 
\textit{TTA}~\cite{shanmugam2020and} uses a standard implementation of TTA that combines the test predictions of a single trained temporal CNN but with the augmented samples by different DA methods. 
The other automatic augmentation methods include $\alpha$-trimmed Augmentation (\textit{$\alpha$-Aug})~\cite{fons2021adaptive}, RandAugment (\textit{RandAug})~\cite{Cubuk_2020}, and SapAugment (\textit{SapAug})~\cite{Hu_2021}.

All of the comparison methods were implemented to use the same hyperparameters and augmentation methods as the proposed method. 
Specifically, the methods use the same number of convolutional layers, fully connected layers, etc., as well as the same DA methods and parameters as the network used in the proposed method. 
The methods that have variable DA parameters (RandAug and SapAug) use a range of $\pm50\%$ for the DA method.

\begin{figure}%[h]
\footnotesize
\centering
\hspace{-5mm}
\begin{tabular}{cc}
\begin{minipage}[t]{0.49\linewidth}
    \centering
    \includegraphics[keepaspectratio, scale=0.1]{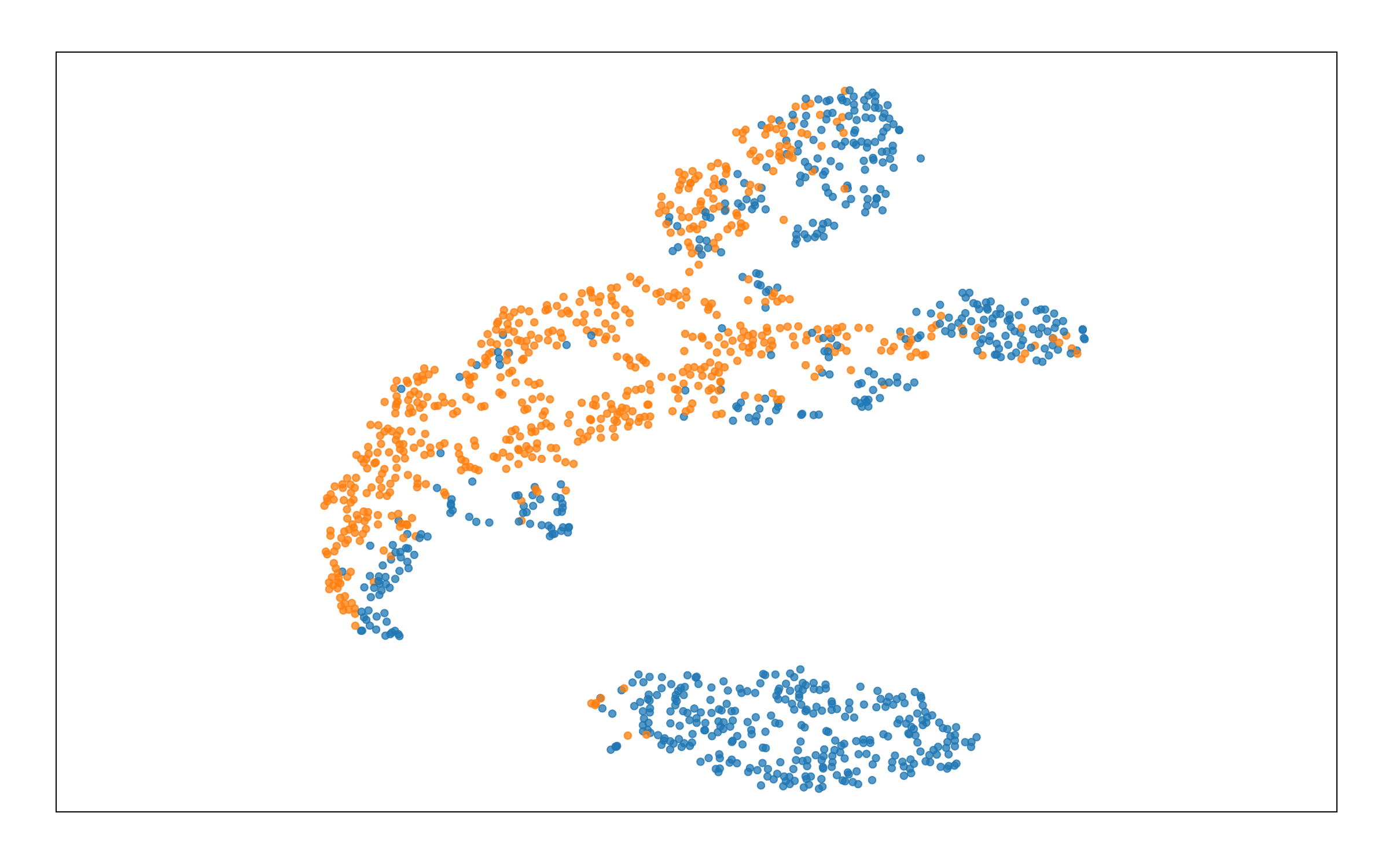}
    {No Augmentation}
\end{minipage}
\begin{minipage}[t]{0.49\linewidth}
    \centering
    \includegraphics[keepaspectratio, scale=0.1]{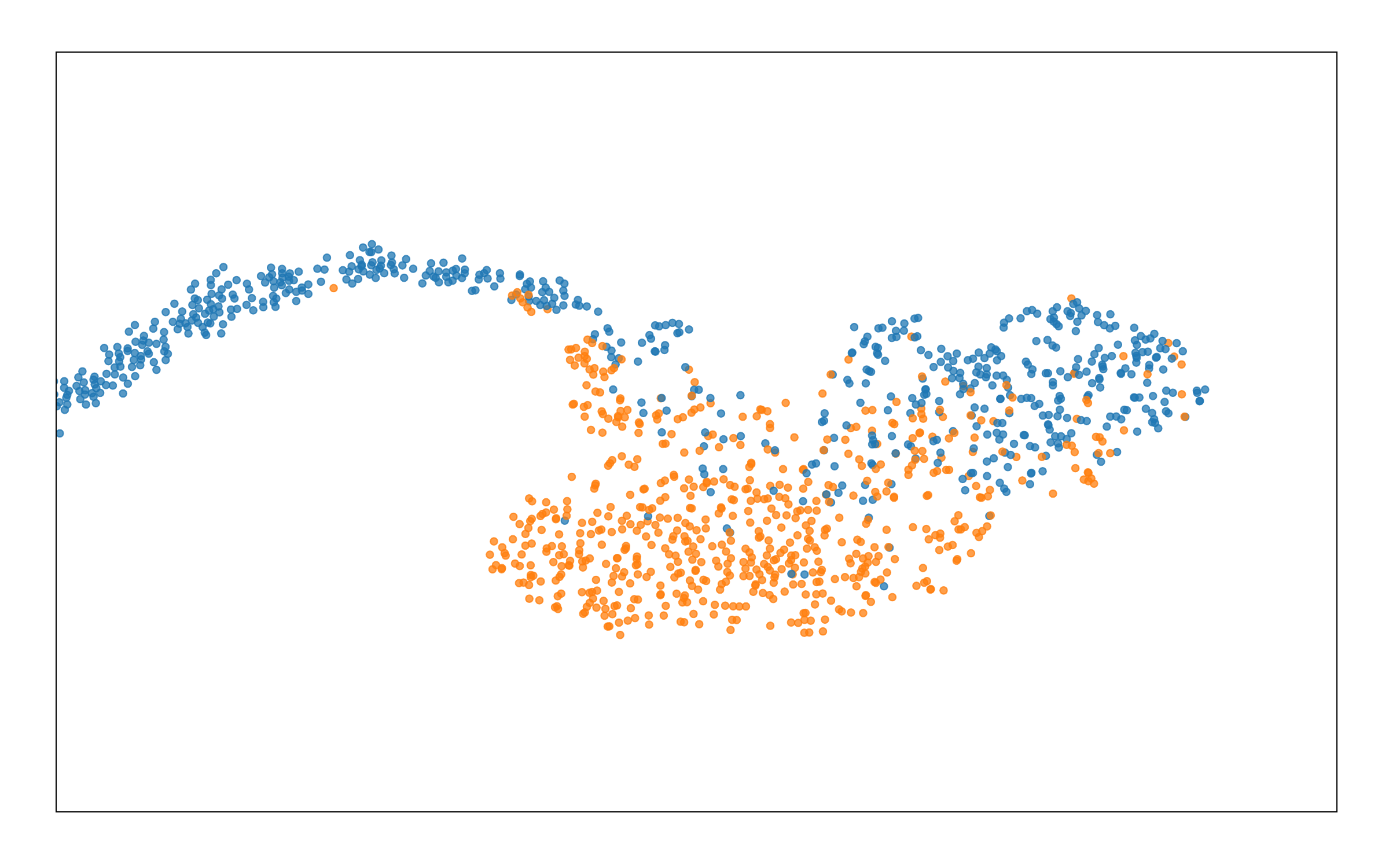}
    {Proposed}
\end{minipage}
\end{tabular}
\caption{Visualization of the combined features $\boldsymbol{f}$ using t-SNE for FordA. The colors correspond to a class. Blue points are class 1 data and orange points are class 2 data.}
\label{fig:tsne}
\end{figure}

\begin{table*}%[!t]
    \renewcommand{\arraystretch}{1.3}
    \caption{Accuracy (\%) Comparison}
    \label{tab:results}
    \centering
    \begin{tabular}{l||ccc|cccc|ccccc}
    \hline
     & \multicolumn{3}{c|}{{Proposed}} & No Aug & w/o GateNet & w/o $\mathcal{L}_\mathrm{con}$ & Shared & TTA & $\alpha$-Aug & RandAug & SapAug \\ 
    Dataset & $\lambda=10.0$ & $1.0$ & $0.1$ & & \& $\mathcal{L}_\mathrm{con}$ & ($\lambda=0.0$) & Weights &  \cite{shanmugam2020and} & \cite{fons2021adaptive} & \cite{Cubuk_2020}  & \cite{Hu_2021} \\ 
    \hline
    Crop  & 75.9 & 76.2 & \textbf{76.8} & 76.1 & 76.1 & 76.1 & 75.8 & 71.5 & 70.7 & 54.5 & 73.0 \\ 
    ElectricDevices & \textbf{68.8} & 68.4 & 67.4 & 68.0 & 68.5 & 66.8 & 66.9 & 65.5 & 64.4 & 61.9 & 67.1 \\ 
    FordA & 89.6 & \textbf{90.2} & 88.8 & 87.2 & 88.8 & 87.2 & 89.1 & 89.2 & 88.6 & 80.1 & 88.4 \\
    FordB & 72.6 & \textbf{72.7} & 71.7 & 70.6 & 71.6 & 70.6 & 68.7 & 71.7 & 70.9 & 60.7 & 71.6 \\  
    HandOutlines & 90.6 & 90.6 & \textbf{91.6} & 90.1 & 91.3 & 91.1 & 80.9 & 45.5 & 90.5 & 64.1 & 91.4 \\ 
    MelbournePed & 94.7 & 95.7 & 90.0 & 95.0 & \textbf{96.0} & 94.5 & 72.2 & 84.7 & 92.7 & 24.4 & 92.3 \\ 
    NonInFECGTx1 & 86.3 & 91.3 & 90.1 & 90.0 & 88.7 & 88.8 & 73.5 & 81.3 & \textbf{92.3} & 5.99 & 87.6 \\ 
    NonInFECGTx2 & 90.5 & 91.8 & 91.7 & 88.2 & 91.7 & 90.5 & 85.1 & 75.8 & \textbf{93.4} & 10.1 & 92.6 \\ 
    PhalangesOC & 79.5 & 82.4 & 82.2 & 80.4 & 82.8 & 82.8 & 74.7 & 53.8 & \textbf{82.9} & 61.3 & 81.9 \\ 
    StarLightCurves & \textbf{97.2} & 96.6 & 95.6 & 95.7 & 96.1 & 96.6 & 88.3 & 94.4 & 95.4 & 63.0 & 96.0 \\ 
    TwoPatterns & 98.0 & \textbf{100} & 99.9 & 95.6 & 99.9 & 99.9 & \textbf{100} & 95.8 & 99.9 & 61.2 & 99.8 \\ 
    Wafer & 99.5 & \textbf{99.6} & \textbf{99.6} & \textbf{99.6} & 99.5 & \textbf{99.6} & 99.3 & 99.5 & 99.5 & 89.2 & \textbf{99.6} \\ 
    \hline
    \end{tabular}
\end{table*}

\subsection{Ablation Study}
To validate the work of introduced modules, we conducted experiments in the three settings.

\textit{No Aug} is the baseline method that uses the same temporal CNN but with no augmentation methods. Its architecture is a combination of a single temporal CNN and the Classifier.

In addition, to observe how our architecture works as a whole, we made a model \textit{without both modules}: the Gating Network and $\mathcal{L}_\mathrm{con}$. In this model, each feature is simply added, or:
\begin{eqnarray*}
    \boldsymbol{f} = \boldsymbol{f}_1 + \boldsymbol{f}_2 + \boldsymbol{f}_3 + \boldsymbol{f}_4 + \boldsymbol{f}_5
\end{eqnarray*}

In \textit{without $\mathcal{L}_\mathrm{con}$}, we made the proposed model without the feature consistency loss to show that introduced the Gating Network alone is insufficient and how the loss helps to achieve better performance.

% Added for rebuttal
Lastly, we made a model \textit{Shared Weights} that shares weights between the Expert Networks. We tried to validate the necessity of individual Expert Networks with this model.

\subsection{Results}

Our quantitative results are shown on Table \ref{tab:results}.
Each result is the mean of five trials for reliability. 
Over trials, they are all trained for the same amount of iterations using the same Adam optimizer. 
We set four different $\lambda$ values, which determine the balance between $\mathcal{L}_\mathrm{con}$ and $\mathcal{L}_\mathrm{CE}$, to get an idea of how the weight of $\mathcal{L}_\mathrm{con}$ affects the performance.

% Competitors
From the table, it is noticeable that ours performed better than the other recent comparative methods in most cases.
In fact, the proposed method ($\lambda=1.0$) achieves better than TTA and RandAug in all 12 datasets. It also outperforms $\alpha$-Aug and SapAug in 9 datasets.

% No Aug
Compared to No Aug, ours performed better on all datasets except HandOutlines and PhalangesOC. Based on this, we can say data augmentation is helpful to achieve better accuracy in general.
To demonstrate the difference in the features of No Aug and the proposed method, we use t-distributed Stochastic Neighbor Embedding (t-SNE)~\cite{vandermaaten08a}.
Fig.~\ref{fig:tsne} visualizes $\boldsymbol{f}$ using t-SNE for the FordA dataset by its class. 
For this visualization, we used the closest model to the mean accuracy of five trials.
As shown in the figure, the proposed method succeeds in creating a representation that is easier to classify than No Aug.
This behavior's effect can be confirmed by the fact that accuracy has improved from $87.2\%$ to $90.2\%$ in Table \ref{tab:results}.

% \subsection{Ablation Results}
% Ablation
Focusing on the model without our two modules shows that it only achieves the best in one dataset while the proposed methods achieve the best around three datasets on average. Moreover, the model with the Gating Network and no $\mathcal{L}_\mathrm{con}$ still achieves the best in one dataset alone. This clearly supports the idea that $\mathcal{L}_\mathrm{con}$ is necessary for effective dynamic data augmentation.

\begin{figure}
\centering
  \begin{minipage}[b]{0.49\linewidth}
    \centering
    \includegraphics[keepaspectratio, scale=0.05]{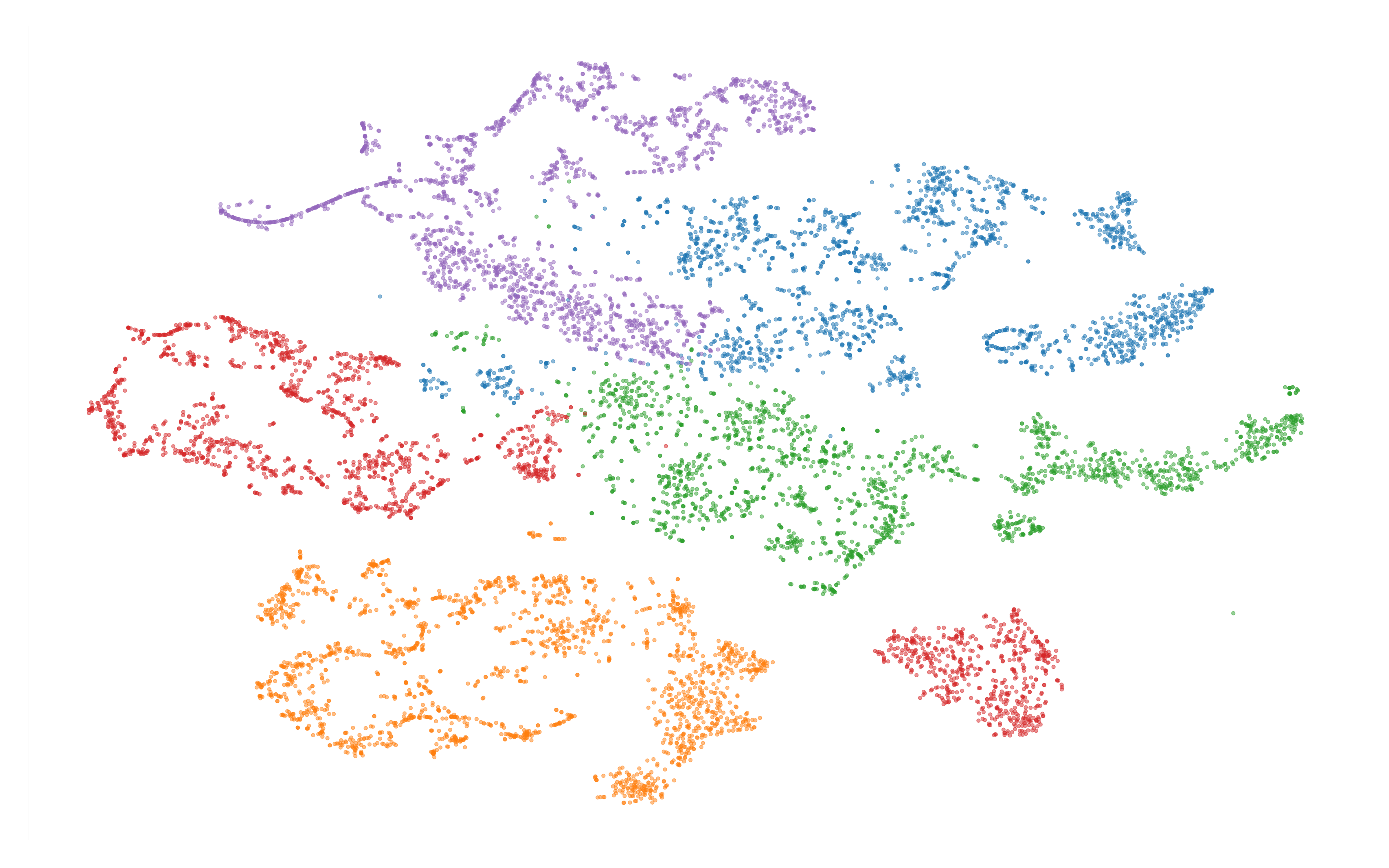}
    {Without $\mathcal{L}_\mathrm{con}$ ($\lambda = 0.0$)}
  \end{minipage}
  \begin{minipage}[b]{0.49\linewidth}
    \centering
    \includegraphics[keepaspectratio, scale=0.05]{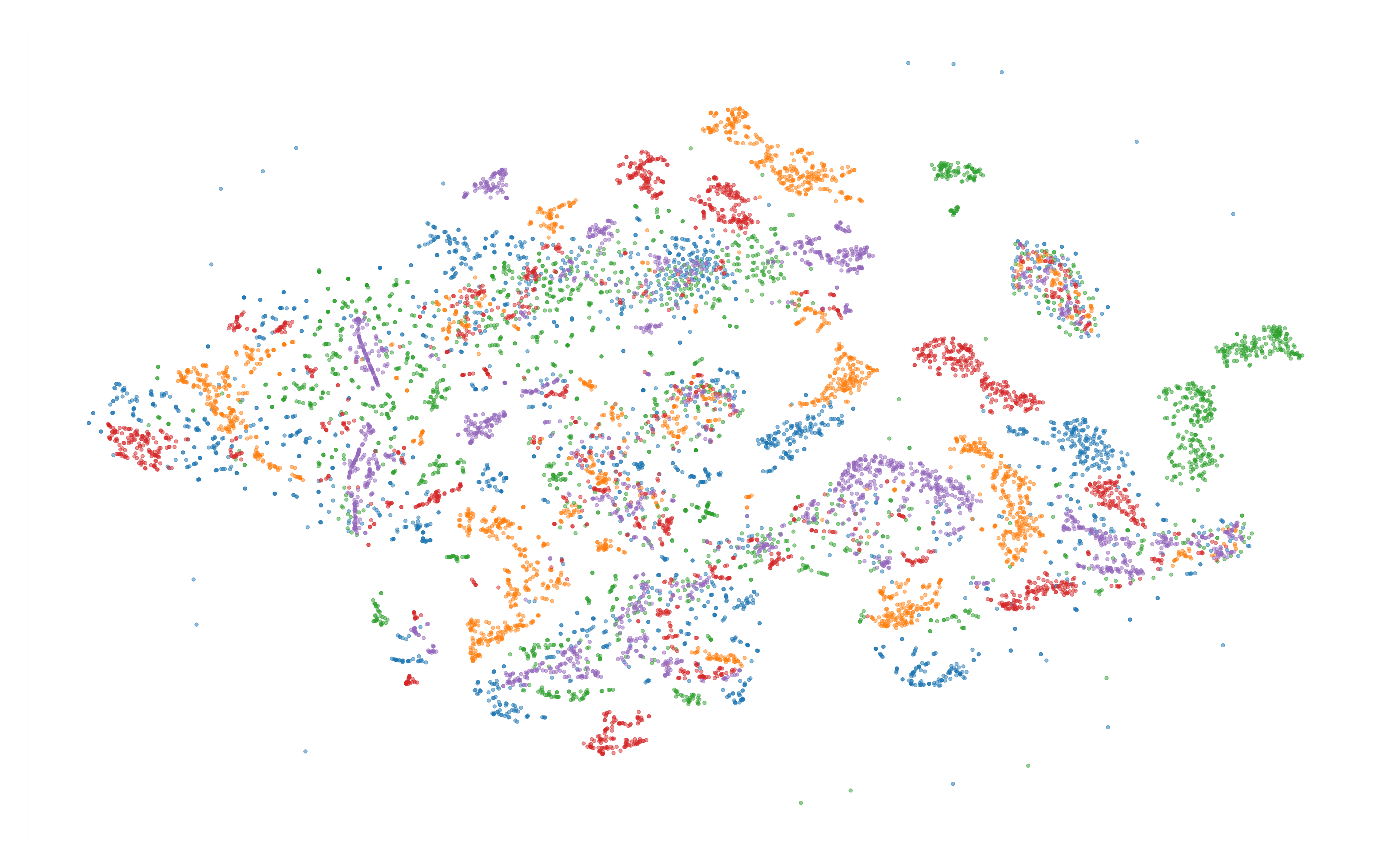}
    {$\lambda = 0.1$}
  \end{minipage} \\
  \begin{minipage}[b]{0.49\linewidth}
    \centering
    \includegraphics[keepaspectratio, scale=0.05]{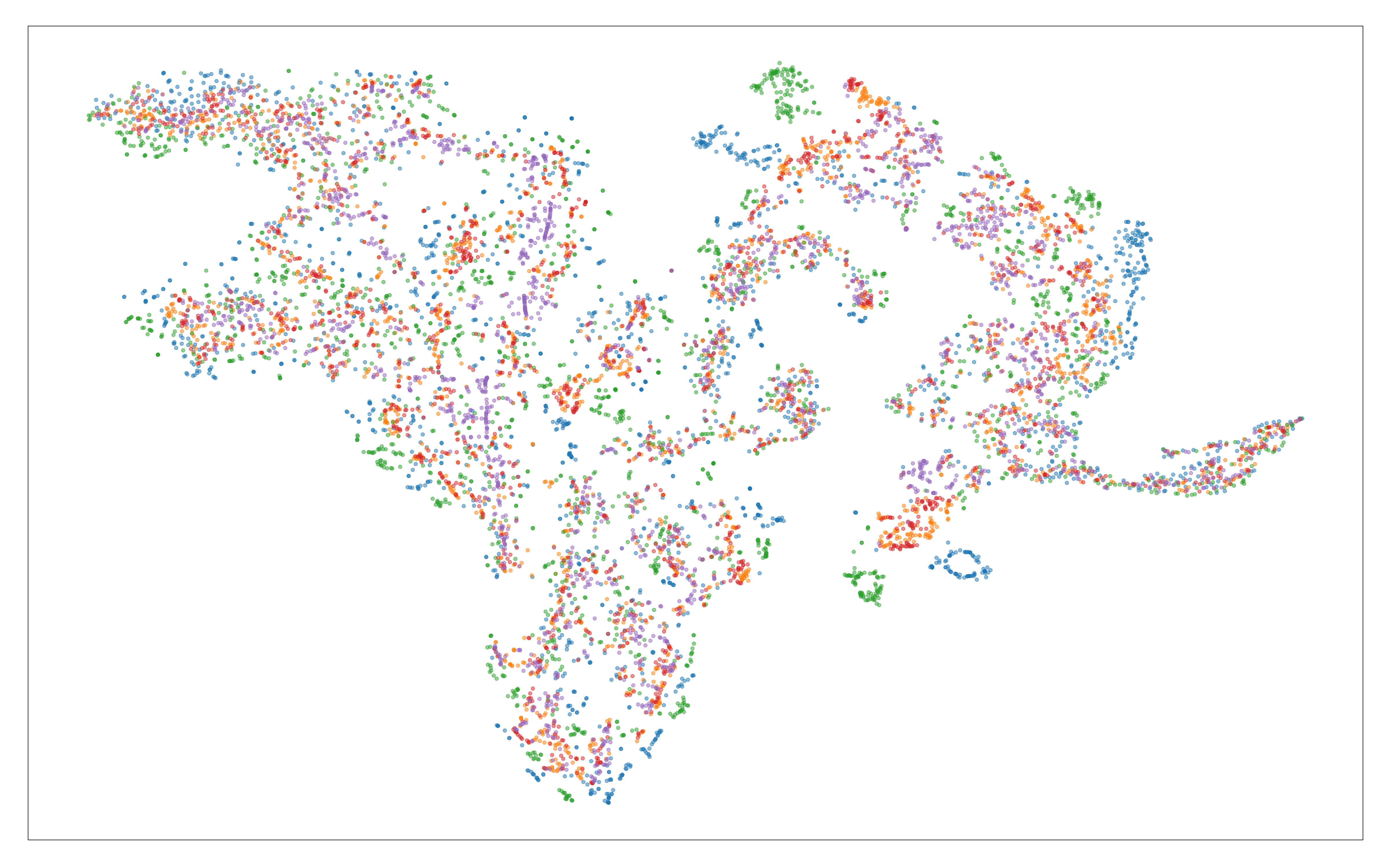}
    {$\lambda = 1.0$}
  \end{minipage}
  \begin{minipage}[b]{0.49\linewidth}
    \centering
    \includegraphics[keepaspectratio, scale=0.05]{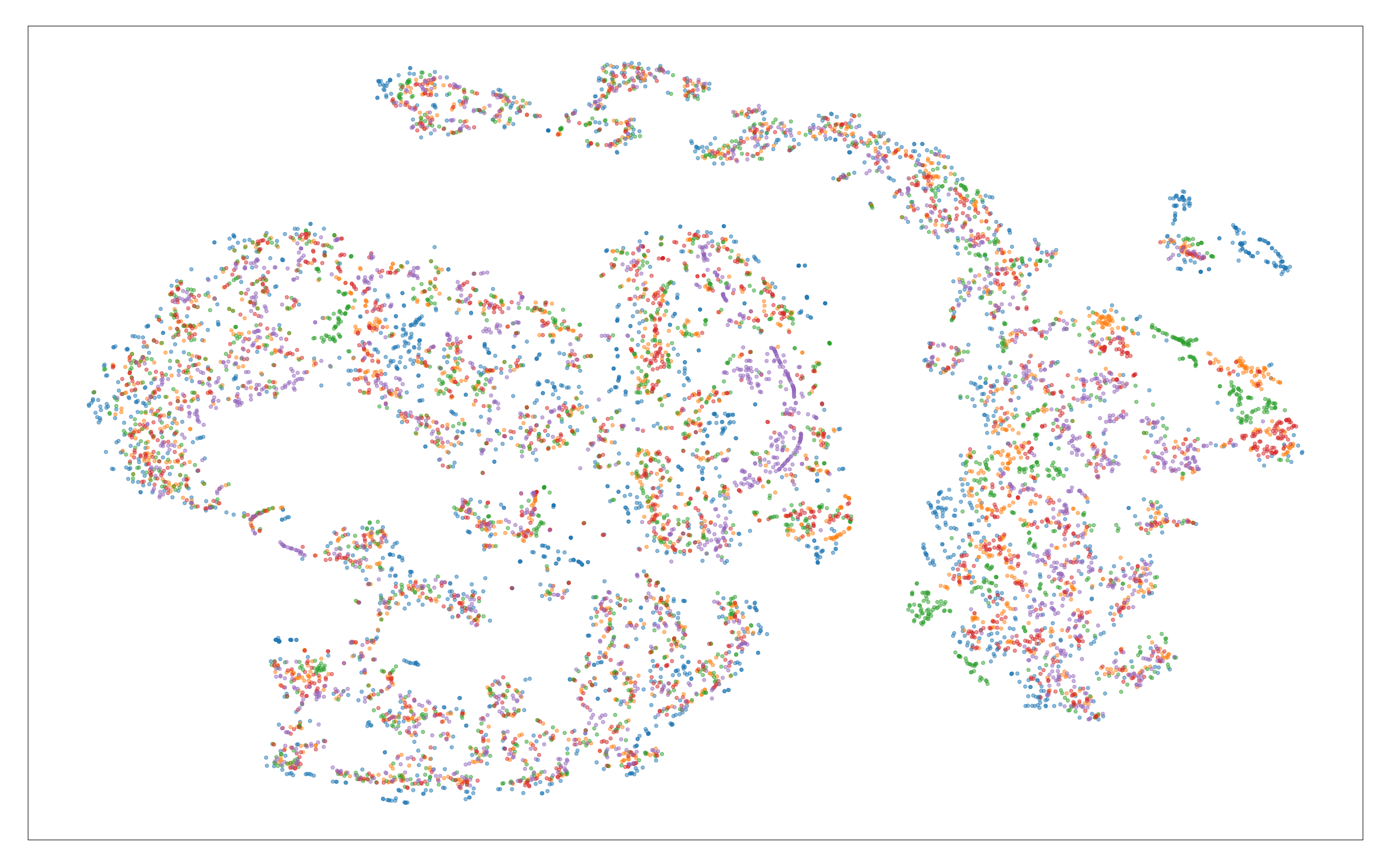}
    {$\lambda = 10.0$}
  \end{minipage}
  \caption{Visualization of the features before the Gating Network using t-SNE for ElectricDevices. Each color represents a different Expert Network. The distribution of dots by color gets more similar as $\lambda$ gets larger.}
  %Identity is the blue points. Jittering is the orange points. Magnitude Warping is green points. Time Warping is the purple points. Window Warping is the red points.}
\label{fig:consisloss}
\vspace{-3mm}
\end{figure}

% FC loss visualize (Fig.5)
We also tried to analyze the work of $\mathcal{L}_\mathrm{con}$ in a qualitative way.
Fig.~\ref{fig:consisloss} is a combined visualization of each $\boldsymbol{f}_n$ for the ElectricDevices dataset colored by DA methods (Expert Network).
Without $\mathcal{L}_\mathrm{con}$, the features are separated based on the DA method. It means when the Gating Network, without $\mathcal{L}_\mathrm{con}$, adds the features, interference and additional class confusion might occur.
However, as $\lambda$ gets bigger, the distribution of dots by color gets closer, allowing the Gating Network to add features in a meaningful way.

% Added for rebuttal
Lastly, the performance of Shared Weights is limited.
It may be concluded that a single Expert Network is not sufficient for multiple DA methods.

\section{Distribution of $\boldsymbol{\alpha}$}

\begin{table}%[!t]
    \renewcommand{\arraystretch}{1.3}
    \caption{Average $\boldsymbol{\alpha}$ ($\lambda=1.0$)}
    \label{tab:alpha}
    \centering
    \begin{tabular}{l|ccccc}
    \hline
     & Ide & Jit & MW & WW & TW \\ 
    Dataset & $\alpha_1$ & $\alpha_2$ & $\alpha_3$ & $\alpha_4$ & $\alpha_5$ \\
    \hline\hline
    %--- result of the model which has median accuracy among 5 ---
    Crop & 0.93 & 0.02 & 0.04 & 0.00 & 0.01 \\ %done
         & $\pm$0.12 & $\pm$0.04 & $\pm$0.06 & $\pm$0.01 & $\pm$0.04 \\
    \hline
    ElectricDevices & 0.72 & 0.00 & 0.27 & 0.00 & 0.00 \\ %done
         & $\pm$0.19 & $\pm$0.00 & $\pm$0.19 & $\pm$0.00 & $\pm$0.00 \\
    \hline
    FordA & 0.92 & 0.02 & 0.03 & 0.01 & 0.01 \\ %done
         & $\pm$0.12 & $\pm$0.04 & $\pm$0.05 & $\pm$0.02 & $\pm$0.02 \\
    \hline
    FordB & 0.86 & 0.06 & 0.06 & 0.02 & 0.01 \\ %done
         & $\pm$0.16 & $\pm$0.09 & $\pm$0.08 & $\pm$0.04 & $\pm$0.02 \\
    \hline
    HandOutlines & 0.99 & 0.00 & 0.00 & 0.00 & 0.00 \\ %done
         & $\pm$0.01 & $\pm$0.00 & $\pm$0.00 & $\pm$0.00 & $\pm$0.00 \\
    \hline
    MelbournePed & 0.91 & 0.06 & 0.02 & 0.01 & 0.00 \\ %done
         & $\pm$0.20 & $\pm$0.16 & $\pm$0.04 & $\pm$0.01 & $\pm$0.01 \\
    \hline
    NonInFetECGTx1 & 0.99 & 0.00 & 0.00 & 0.00 & 0.00 \\ %done
         & $\pm$0.01 & $\pm$0.00 & $\pm$0.01 & $\pm$0.00 & $\pm$0.00 \\
    \hline
    NonInFetECGTx2 & 0.83 & 0.00 & 0.17 & 0.00 & 0.00 \\ %done
         & $\pm$0.32 & $\pm$0.01 & $\pm$0.32 & $\pm$0.01 & $\pm$0.00 \\
    \hline
    PhalangesOC & 1.00 & 0.00 & 0.00 & 0.00 & 0.00 \\ %done
         & $\pm$0.00 & $\pm$0.00 & $\pm$0.00 & $\pm$0.00 & $\pm$0.00 \\
    \hline
    StarLightCurves & 0.75 & 0.06 & 0.08 & 0.09 & 0.02 \\ %done
         & $\pm$0.37 & $\pm$0.09 & $\pm$0.13 & $\pm$0.14 & $\pm$0.04 \\
    \hline
    TwoPatterns & 0.65 & 0.17 & 0.05 & 0.12 & 0.01 \\ %done
         & $\pm$0.25 & $\pm$0.13 & $\pm$0.03 & $\pm$0.13 & $\pm$0.01 \\
    \hline
    Wafer & 0.89 & 0.04 & 0.04 & 0.01 & 0.01\\ %done
         & $\pm$0.12 & $\pm$0.04 & $\pm$0.05 & $\pm$0.01 & $\pm$0.01 \\
    \hline
    \end{tabular}
\end{table}

By examining the learned $\boldsymbol{\alpha}$, it is possible to gain insight into the usefulness of the different DA methods. 
Table~\ref{tab:alpha} shows the average $\boldsymbol{\alpha}$ with a standard deviation for the proposed method with $\lambda=1.0$.
%We used the model which achieves the median accuracy over five trials.

\begin{figure}%[h]
\footnotesize
\centering
\begin{tabular}{ccccc}
\begin{minipage}[t]{\linewidth}
    \centering
    \includegraphics[keepaspectratio, scale=0.13]{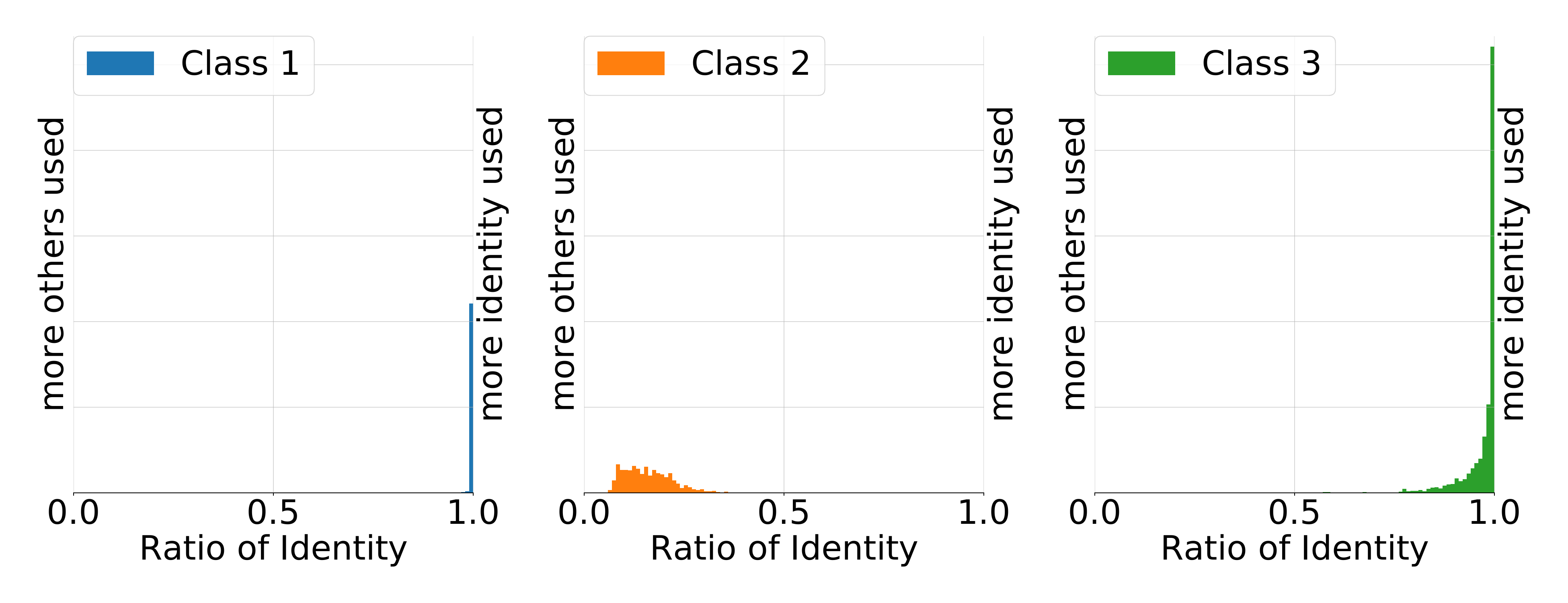}{}
\end{minipage} \\
\begin{minipage}[t]{\linewidth}
    \centering
    \includegraphics[keepaspectratio, scale=0.46]{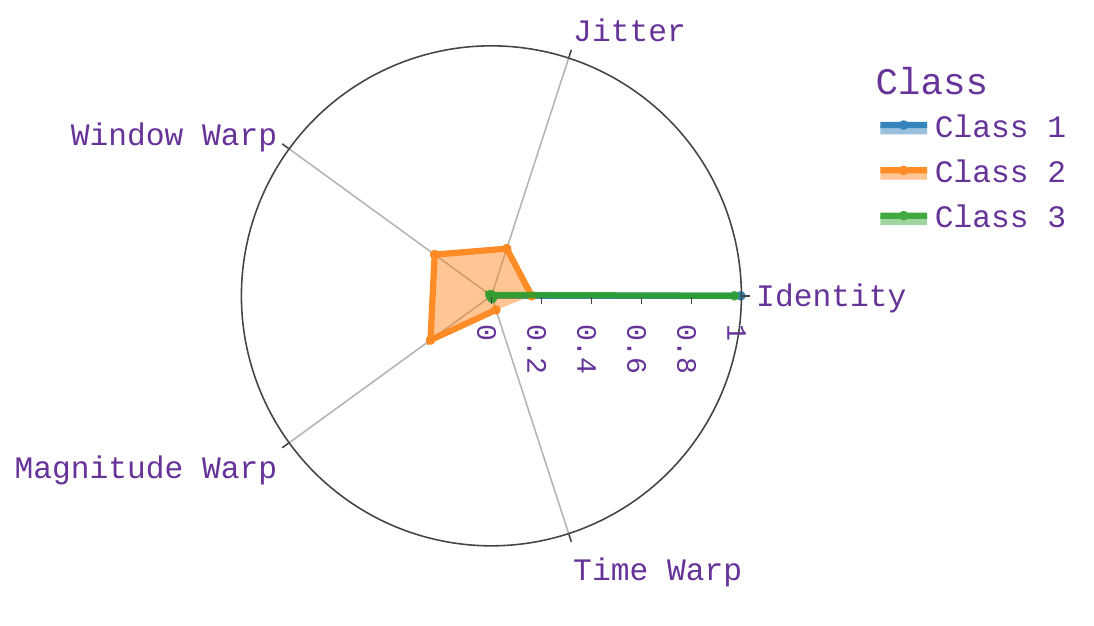}
    {}
\end{minipage} \\
\begin{minipage}[t]{0.187\linewidth}
    \centering
    \includegraphics[keepaspectratio, scale=0.11]{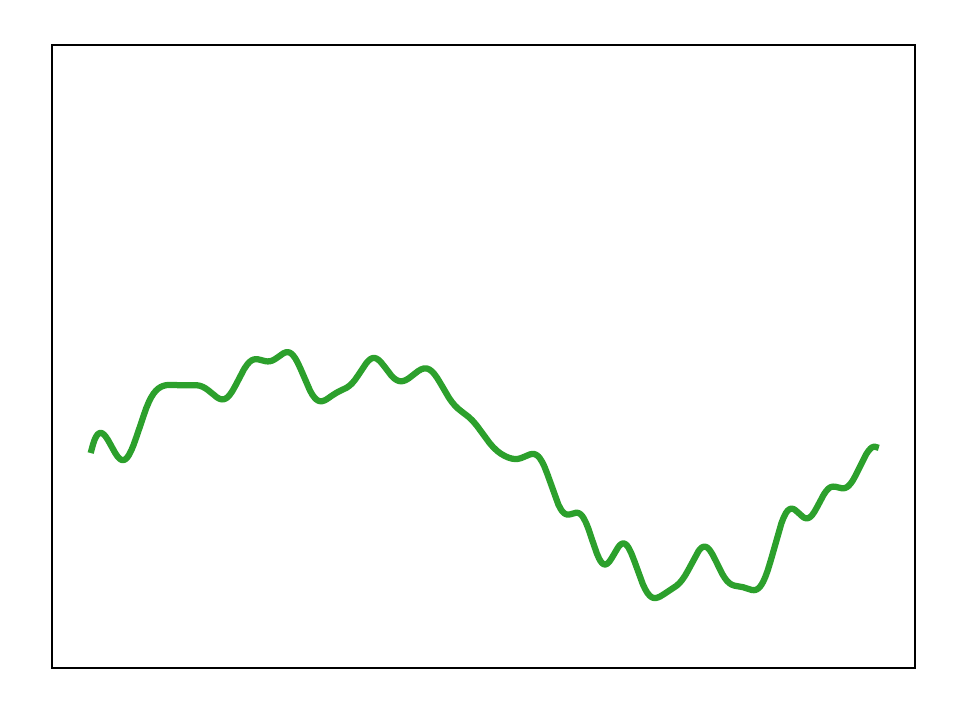}
    {$\alpha_1 = 1.000$}
\end{minipage}
\begin{minipage}[t]{0.187\linewidth}
    \centering
    \includegraphics[keepaspectratio, scale=0.11]{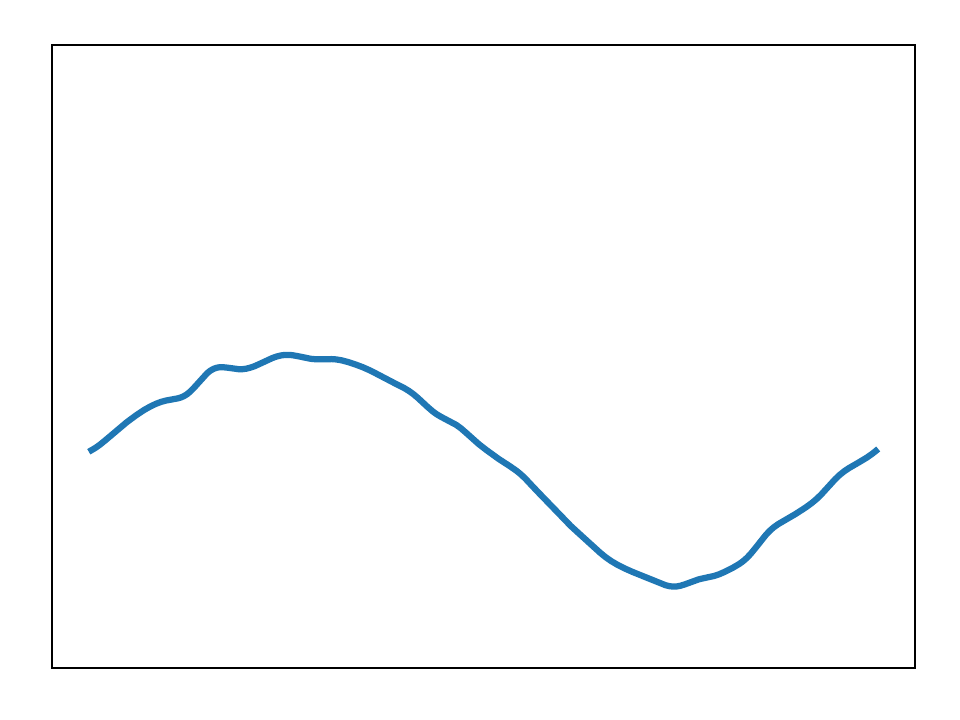}
    {$\alpha_1 = 1.000$}
\end{minipage}
\begin{minipage}[t]{0.187\linewidth}
    \centering
    \includegraphics[keepaspectratio, scale=0.11]{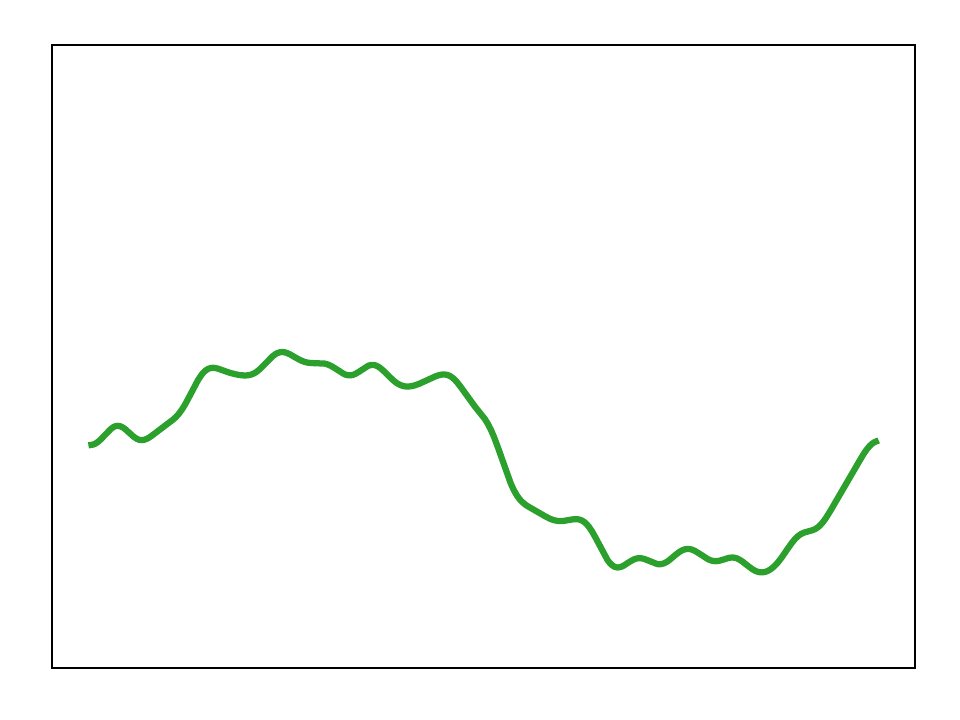}
    {$\alpha_1 = 1.000$}
\end{minipage}
\begin{minipage}[t]{0.187\linewidth}
    \centering
    \includegraphics[keepaspectratio, scale=0.11]{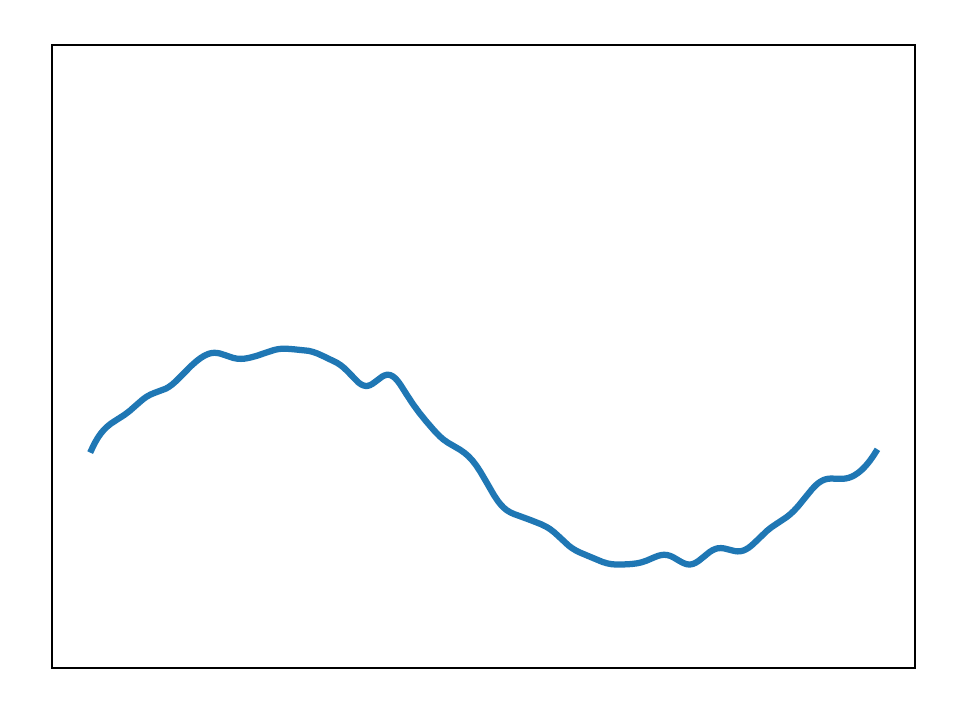}
    {$\alpha_1 = 1.000$}
\end{minipage}
\begin{minipage}[t]{0.187\linewidth}
    \centering
    \includegraphics[keepaspectratio, scale=0.11]{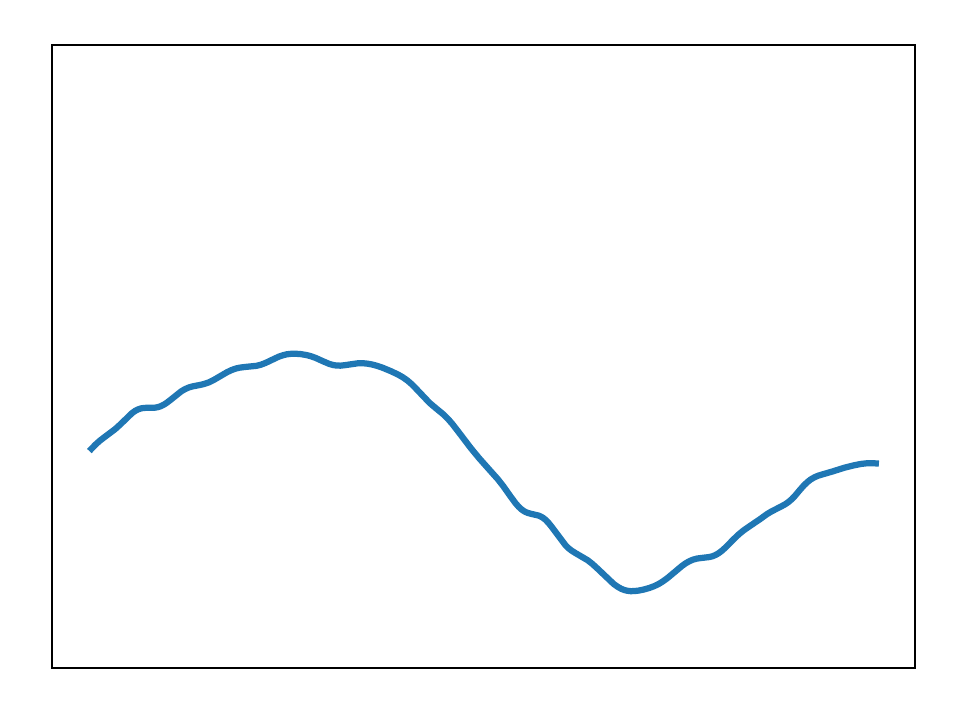}
    {$\alpha_1 = 1.000$}
\end{minipage} \\
\begin{minipage}[t]{0.187\linewidth}
    \centering
    \includegraphics[keepaspectratio, scale=0.11]{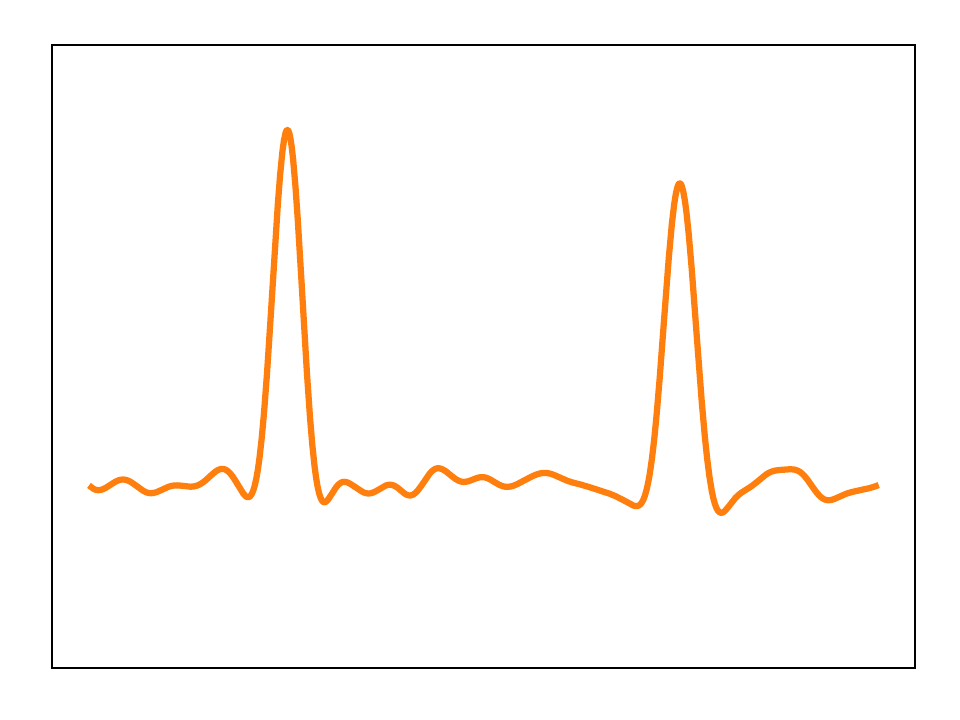}
    {$\alpha_1 = 0.059$}
\end{minipage}
\begin{minipage}[t]{0.187\linewidth}
    \centering
    \includegraphics[keepaspectratio, scale=0.11]{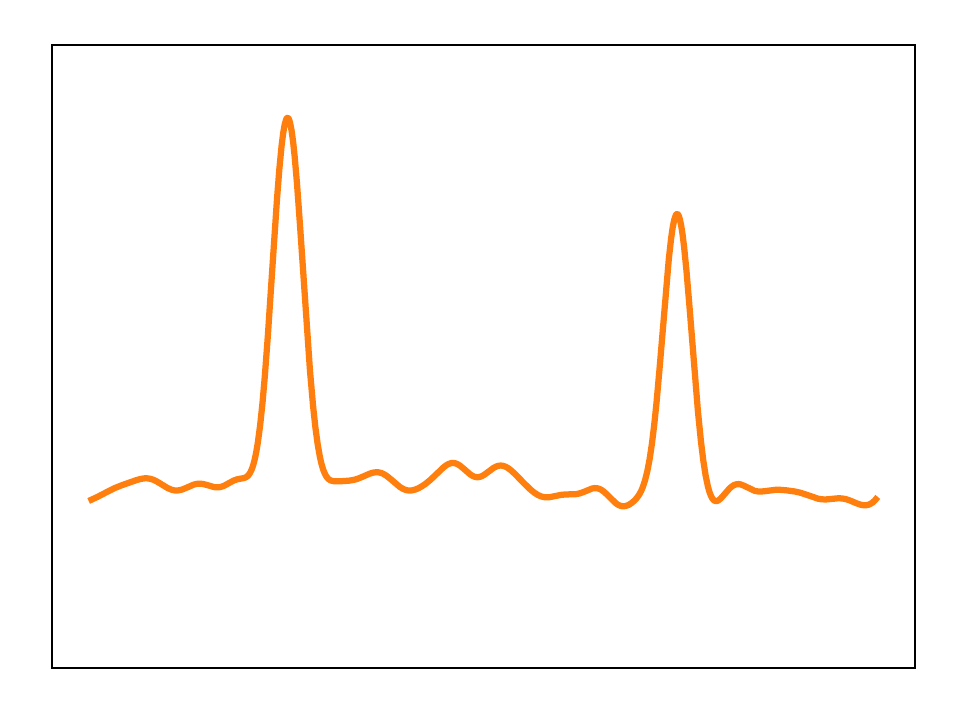}
    {$\alpha_1 = 0.064$}
\end{minipage}
\begin{minipage}[t]{0.187\linewidth}
    \centering
    \includegraphics[keepaspectratio, scale=0.11]{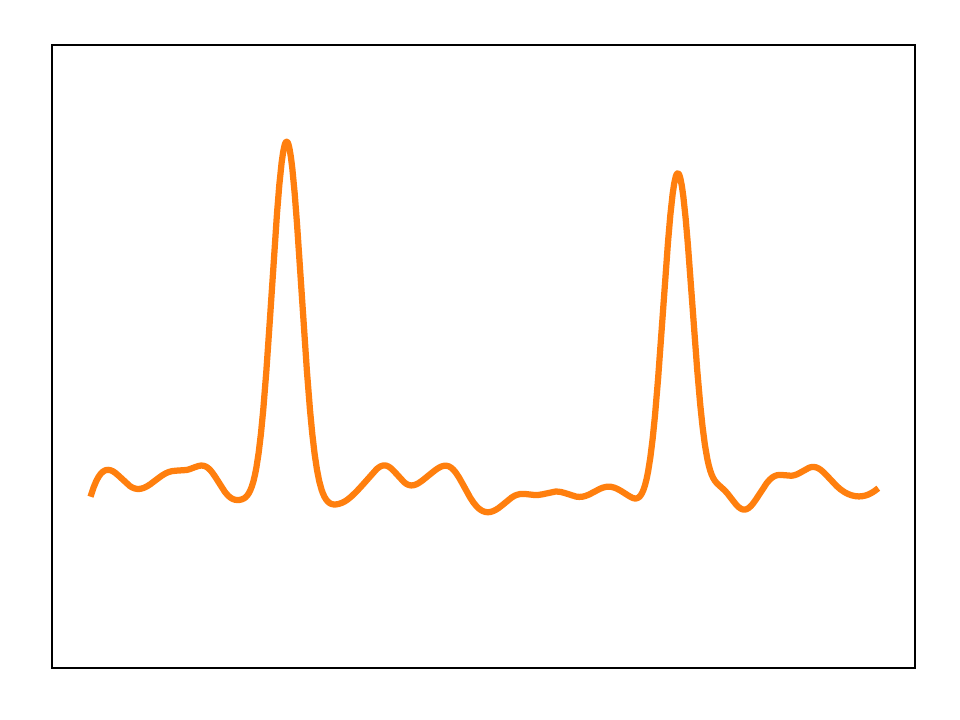}
    {$\alpha_1 = 0.066$}
\end{minipage}
\begin{minipage}[t]{0.187\linewidth}
    \centering
    \includegraphics[keepaspectratio, scale=0.11]{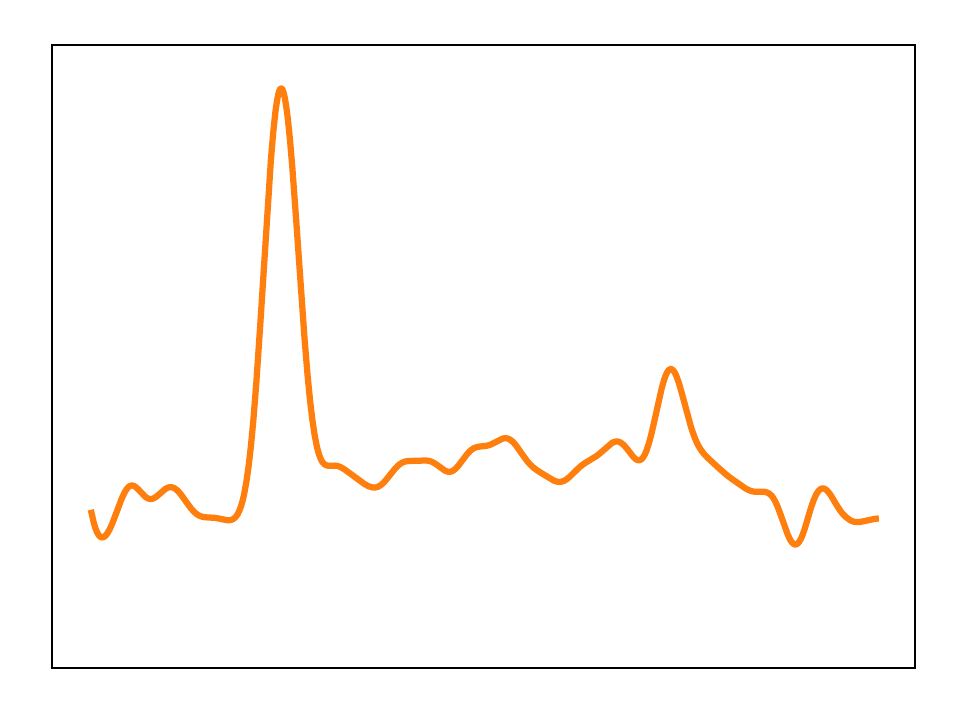}
    {$\alpha_1 = 0.066$}
\end{minipage}
\begin{minipage}[t]{0.187\linewidth}
    \centering
    \includegraphics[keepaspectratio, scale=0.11]{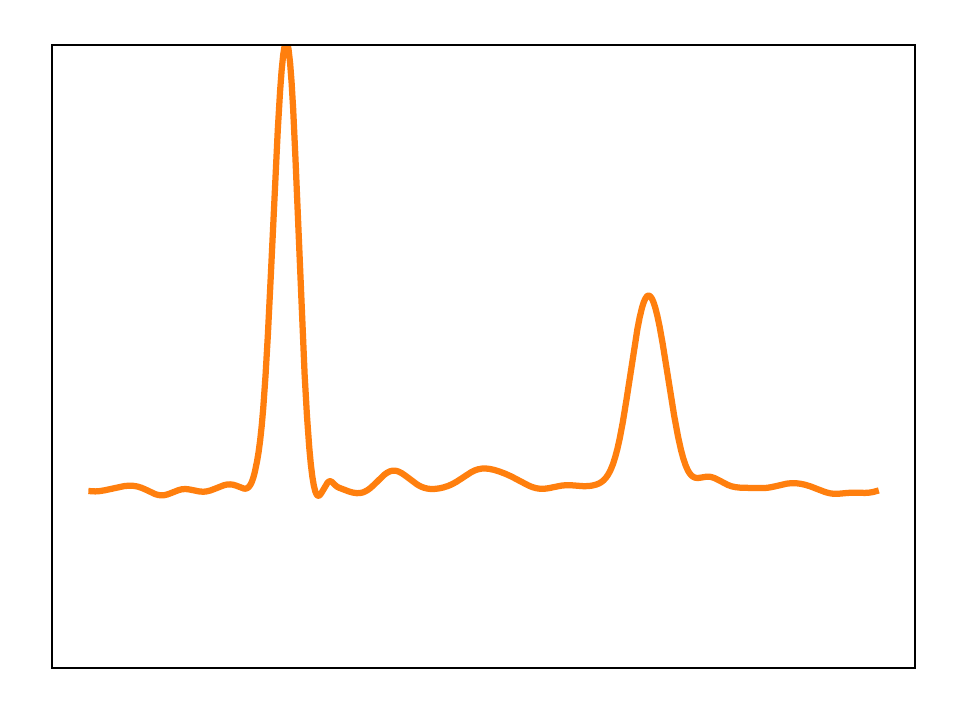}
    {$\alpha_1 = 0.066$}
\end{minipage}
\end{tabular}
\caption{(top) histograms of $\alpha_{identity}$ for StarLightCurves. The x-axis is the ratio between Identity on the right and the others on the left. (middle) radar chart of the ratio of the five DA methods in StarLightCurves. (bottom) five highest and lowest samples of $\alpha_\mathrm{identity}$.}
%\vspace{-21mm}
\label{fig:starlightcurves}
\end{figure}

In general, Identity is used the most over the datasets.
However, this does not imply that the other DA methods are not used, considering their standard deviations.
% , most of the distributions are hill-like shapes, not spike-like shapes. 
This means our model gives non-Identity dependent $\boldsymbol{\alpha}$ on samples that have characteristic shapes.
We visualized the distribution of $\boldsymbol{\alpha}$ to see its relation between $\boldsymbol{\alpha}$ and samples.
Histograms in Fig.\ref{fig:starlightcurves} show the distributions of the ratio of $\alpha$ for Identity versus the others. Samples on the right use features by Identity, and samples on the left use features by the other methods. This result is remarkable because class 2 has a low ratio of $\alpha$ for identity, although class 1 and class 3 congregates at $\alpha_{identity}=1.0$.
From a radar chart in the middle on Fig.\ref{fig:starlightcurves}, the proposed model gives a combination of different DA methods for samples in class 2.
It indicates that for some datasets, classification can be improved by combining specific DA methods.
Fig.~\ref{fig:starlightcurves} bottom shows the five highest and lowest $\alpha_{identity}$ samples on the dataset. We can observe a clear trend that samples whose shapes are similar have similar $\alpha$ values. Although green and blue are different classes, the same $\alpha$ values are given as their shapes are similar. On the other hand, the orange samples, whose shapes are quite different from the two, are given $\alpha$ values that are also quite different from the other classes.

\begin{figure}%[h]
\footnotesize
\centering
\begin{tabular}{ccccc}
\begin{minipage}[t]{\linewidth}
    \centering
    \includegraphics[keepaspectratio, scale=0.13]{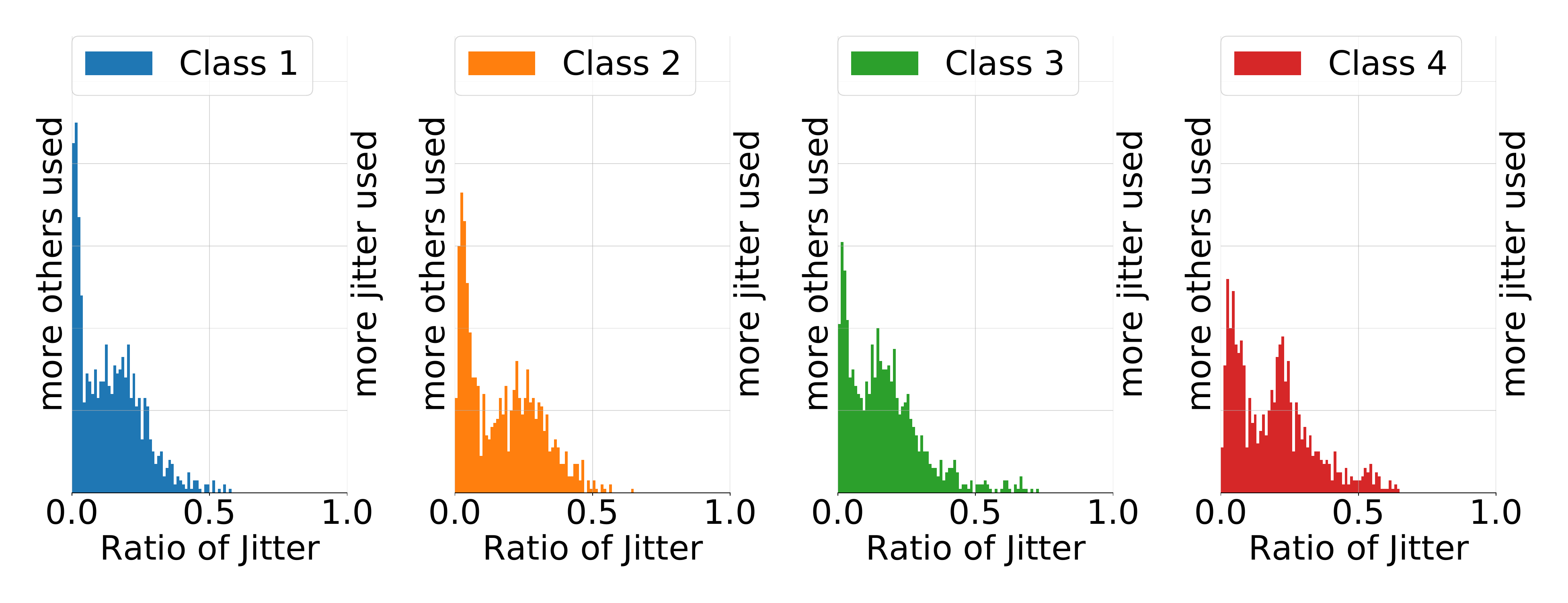}{}
\end{minipage} \\
\begin{minipage}[t]{0.187\linewidth}
    \centering
    \includegraphics[keepaspectratio, scale=0.11]{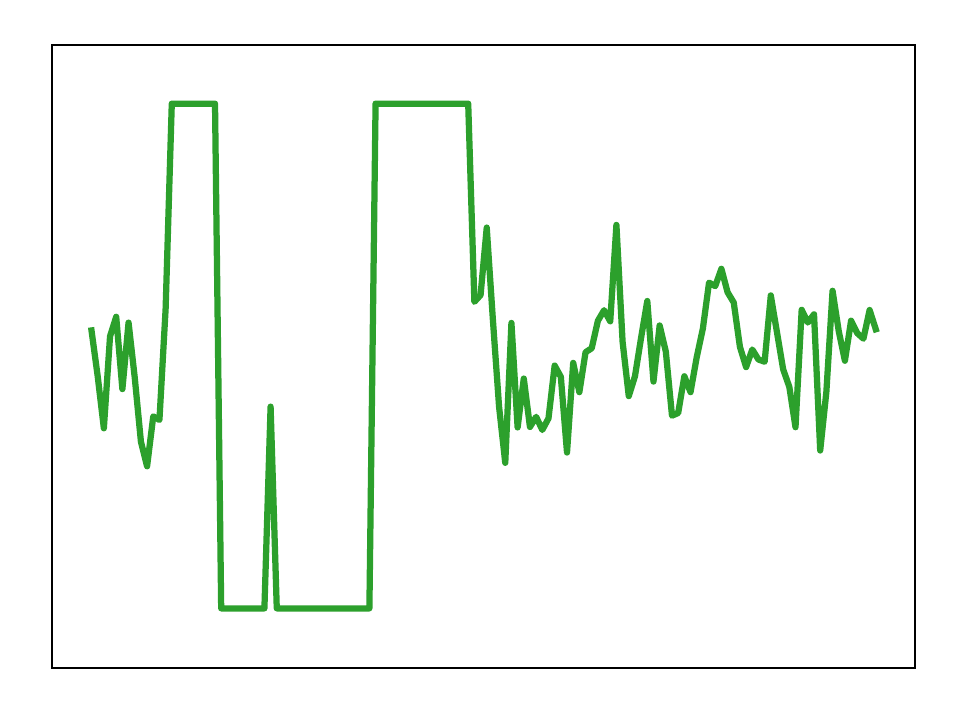}
    {$\alpha_2 = 0.667$}
\end{minipage}
\begin{minipage}[t]{0.187\linewidth}
    \centering
    \includegraphics[keepaspectratio, scale=0.11]{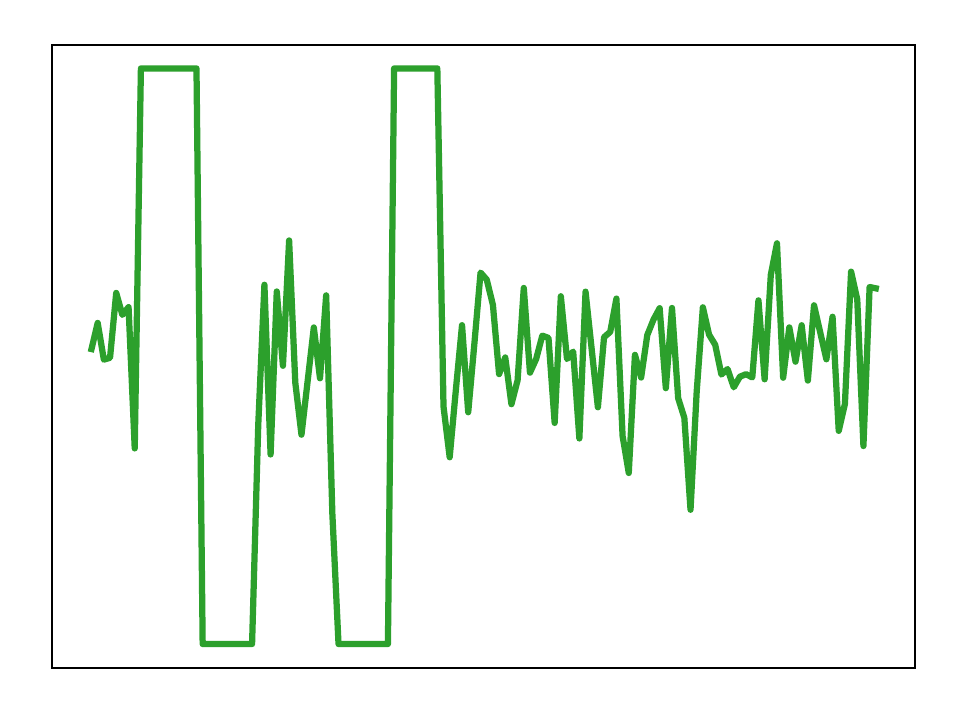}
    {$\alpha_2 = 0.677$}
\end{minipage}
\begin{minipage}[t]{0.187\linewidth}
    \centering
    \includegraphics[keepaspectratio, scale=0.11]{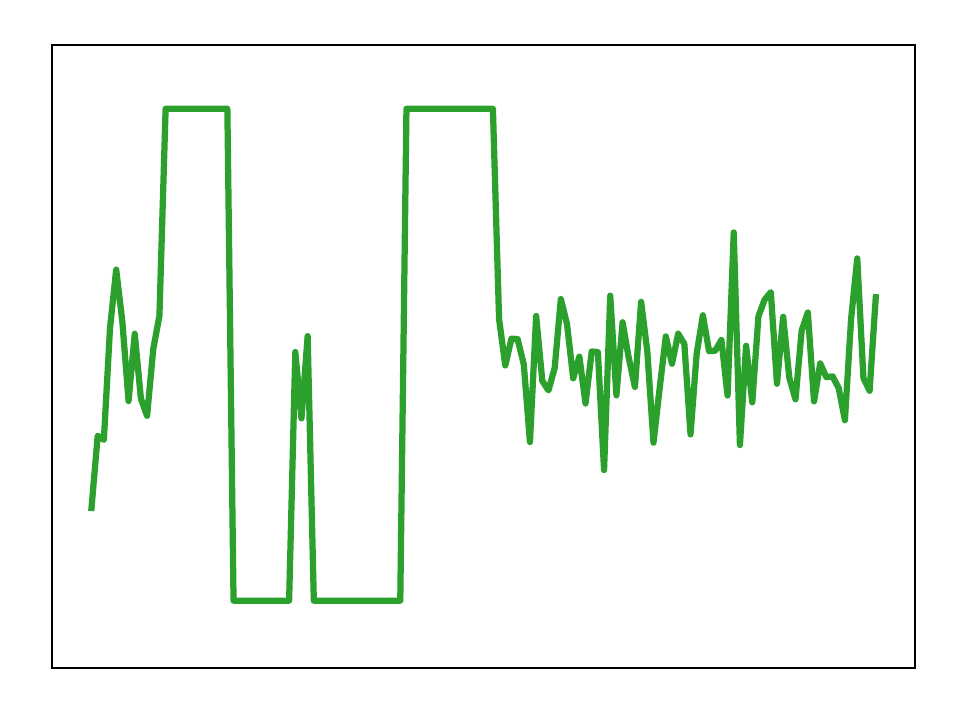}
    {$\alpha_2 = 0.685$}
\end{minipage}
\begin{minipage}[t]{0.187\linewidth}
    \centering
    \includegraphics[keepaspectratio, scale=0.11]{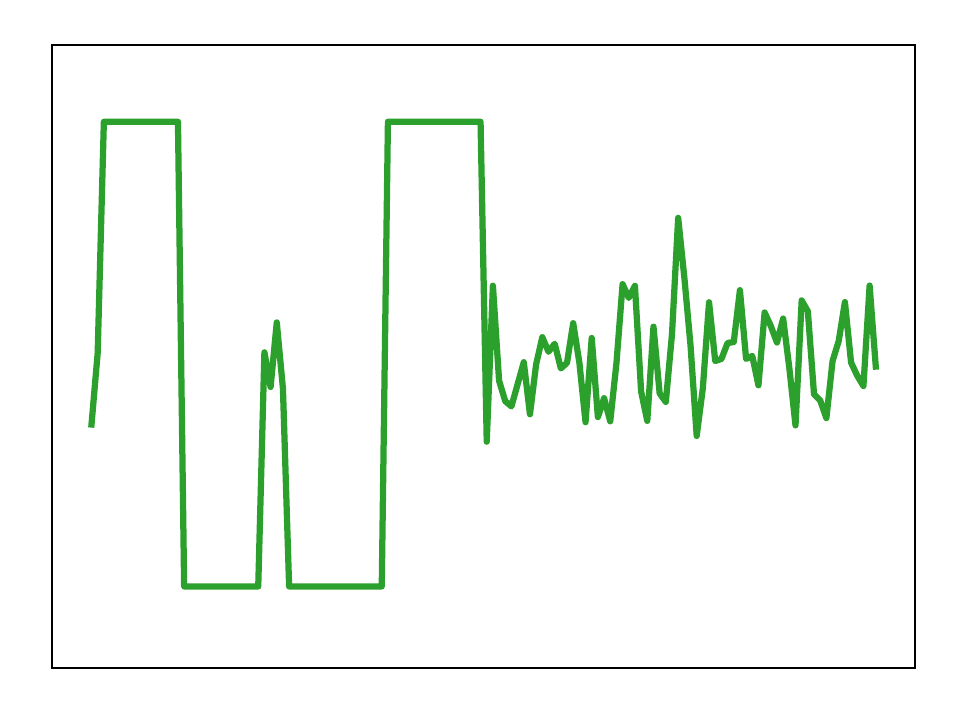}
    {$\alpha_2 = 0.705$}
\end{minipage}
\begin{minipage}[t]{0.187\linewidth}
    \centering
    \includegraphics[keepaspectratio, scale=0.11]{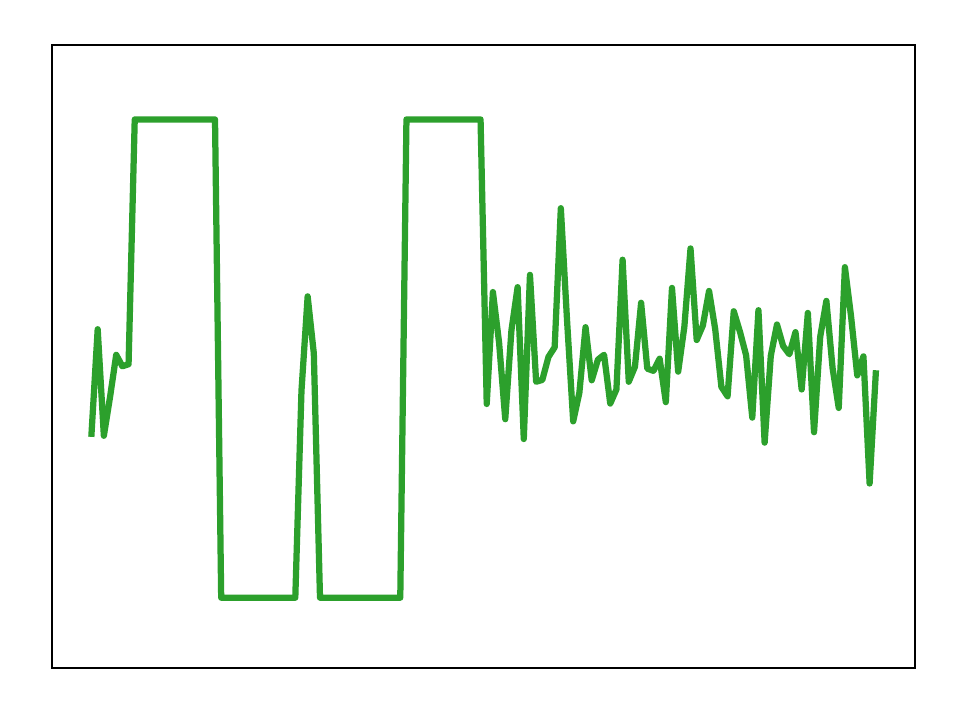}
    {$\alpha_2 = 0.724$}
\end{minipage} \\
\begin{minipage}[t]{0.187\linewidth}
    \centering
    \includegraphics[keepaspectratio, scale=0.11]{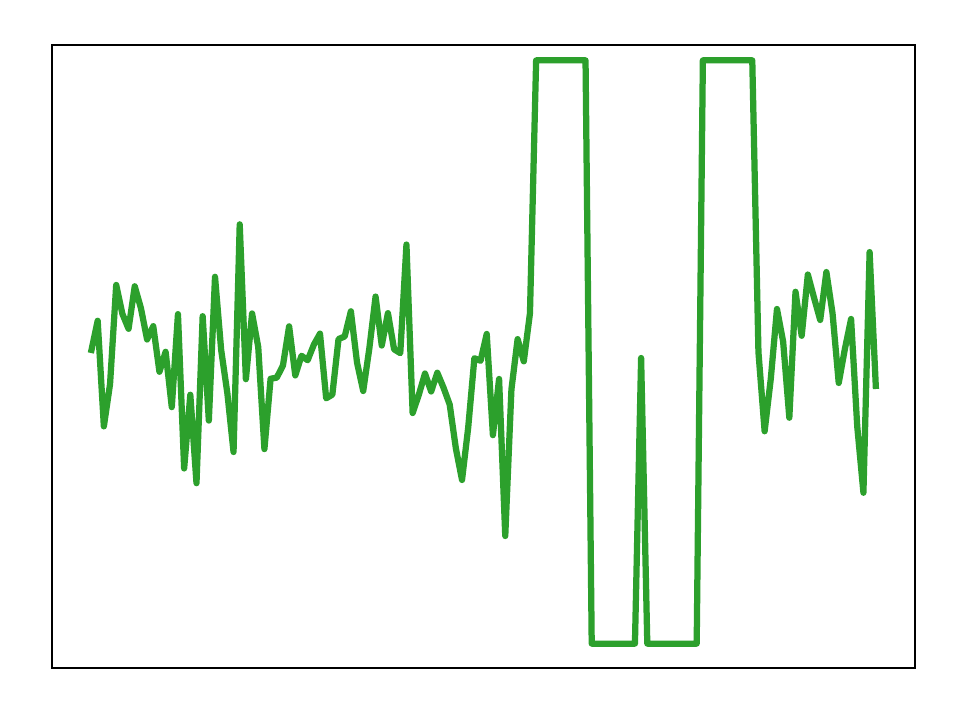}
    {$\alpha_2 = 0.001$}
\end{minipage}
\begin{minipage}[t]{0.187\linewidth}
    \centering
    \includegraphics[keepaspectratio, scale=0.11]{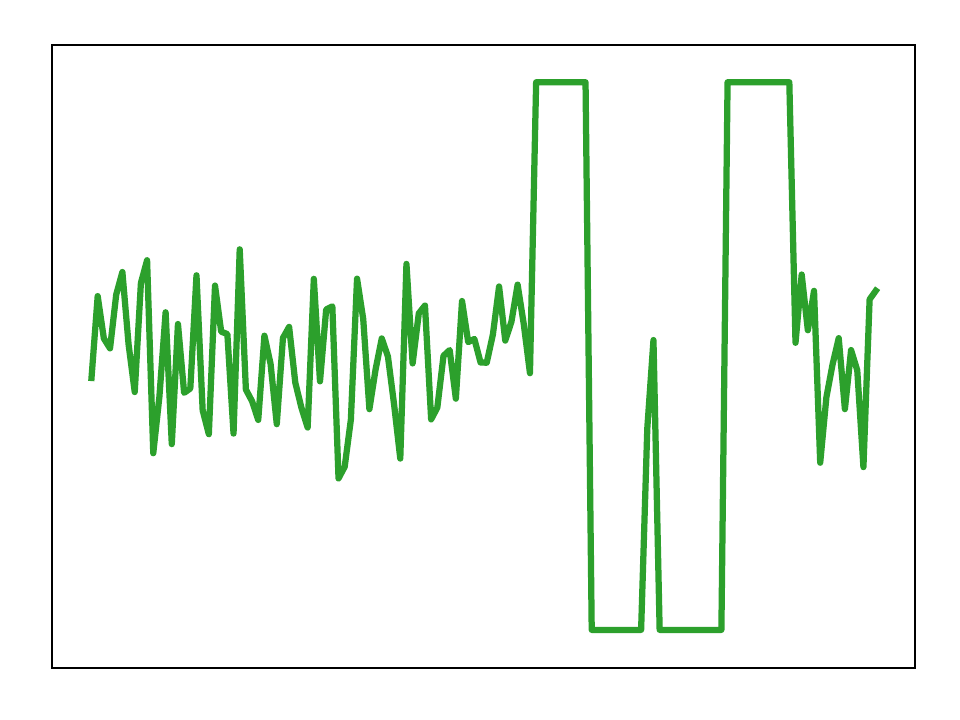}
    {$\alpha_2 = 0.001$}
\end{minipage}
\begin{minipage}[t]{0.187\linewidth}
    \centering
    \includegraphics[keepaspectratio, scale=0.11]{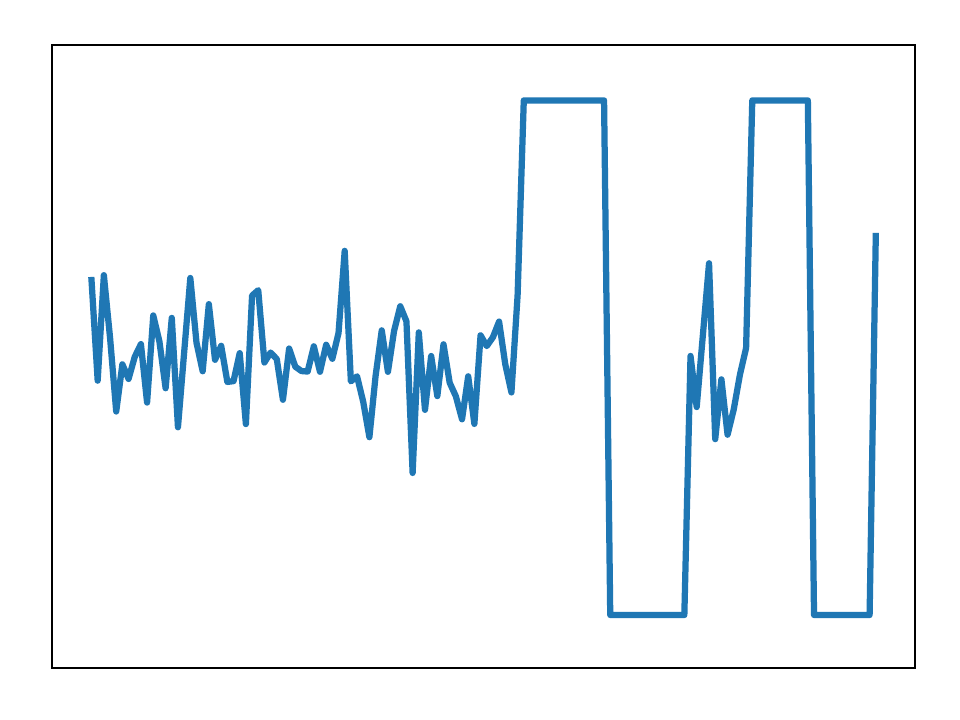}
    {$\alpha_2 = 0.001$}
\end{minipage}
\begin{minipage}[t]{0.187\linewidth}
    \centering
    \includegraphics[keepaspectratio, scale=0.11]{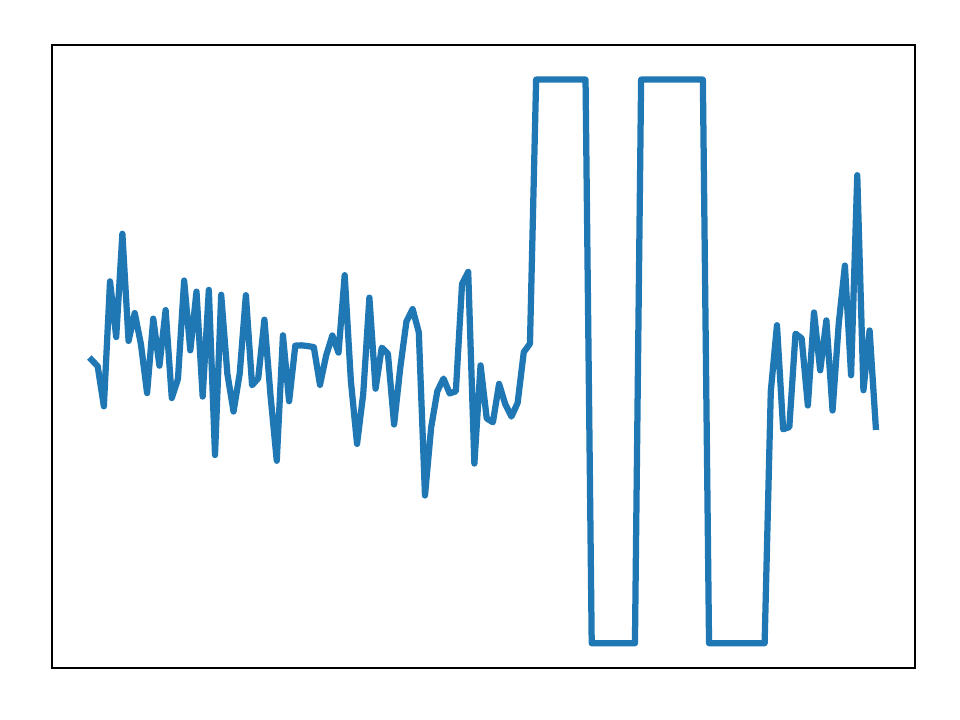}
    {$\alpha_2 = 0.002$}
\end{minipage}
\begin{minipage}[t]{0.187\linewidth}
    \centering
    \includegraphics[keepaspectratio, scale=0.11]{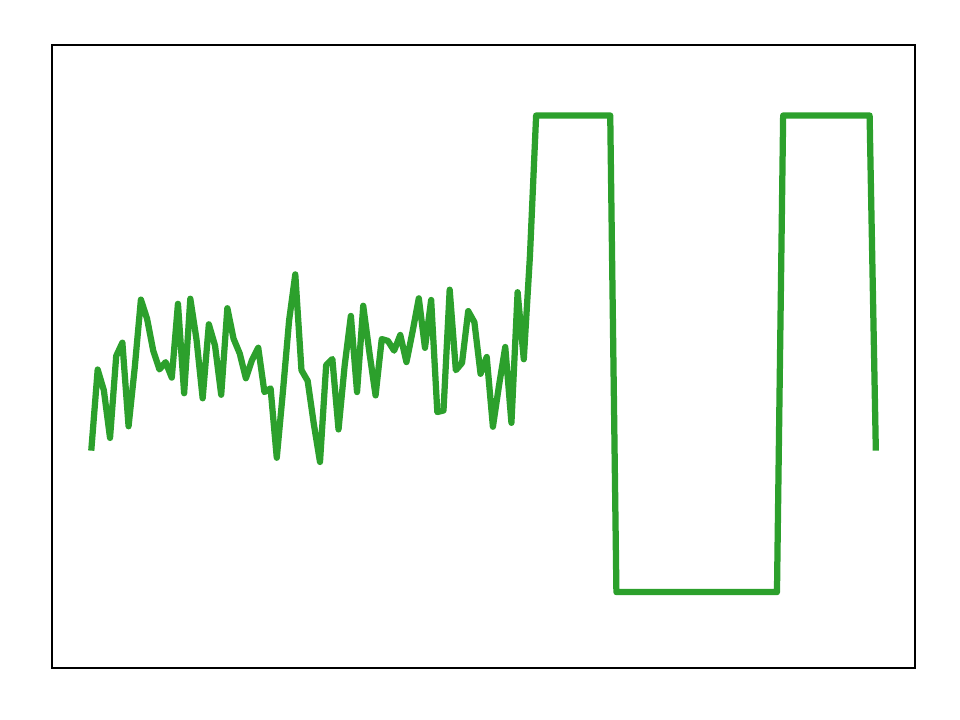}
    {$\alpha_2 = 0.002$}
\end{minipage}
\end{tabular}
\caption{(top) histograms of $\alpha_{jitter}$ for TwoPatterns. The x-axis is the ratio between Jittering on the right and the others on the left. (bottom) five highest and lowest samples of $\alpha_\mathrm{jitter}$.}
\label{fig:twopattern}
\end{figure}

Other interesting observations in Table~\ref{tab:alpha} is that ElectricDevices and NonInFetECGTx2 have relatively higher $\boldsymbol{\alpha}$ values for Magnitude Warping, and StarLightCurves and TwoPatterns rely on Window Warping more.
For Jittering, TwoPattens utilize the method heavily.
Fig.~\ref{fig:twopattern} shows the distribution of $\alpha_{jitter}$ and the five highest and lowest $\alpha_{jitter}$ samples on TwoPatterns.
From the histogram, we can see there are some differences among classes.
For example, class 4 has a steep valley of around 0.15, while class 1 has no valley. In addition, class 3 has a longer right side tail, which ends at about 0.7.
On the list of the highest and lowest samples, class 3 samples are listed on both sides. It implies that the network can give significantly different $\alpha$ values even in the same class.

The last thing to add is that Time Warping consistently has a low $\boldsymbol{\alpha}$. 
This may be attributed to the fact that it is possible for Time Warping to add excessive changes to the original time series.  
Due to this, the model learns not to utilize this method as much as the other ones. 

\section{Conclusion}
\label{sec:conclusion}

In the paper, we proposed a novel neural network that dynamically selects the best combinations of features created by DA methods. 
The proposed method combines multiple Expert Networks with a Gating Network. 
In addition, we use a novel feature consistency loss that ensures that the features used in the Gating Network are similar for each sample.
The proposed method's effectiveness is validated in experiments comparing baseline and other recent automatic DA methods in 12 datasets.
Not only does our model achieves better accuracy than others in most cases, but ours is also explainable through the analysis of the learned weight $\boldsymbol{\alpha}$.

\section*{Acknowledgment}
This work was partially supported by MEXT-Japan
(Grant No. J21K17808) and R3QR Program (Qdai-jump Research Program) 01252.

\newpage


\bibliographystyle{IEEEtran}
\bibliography{da}

\begin{thebibliography}{10}
\providecommand{\url}[1]{#1}
\csname url@samestyle\endcsname
\providecommand{\newblock}{\relax}
\providecommand{\bibinfo}[2]{#2}
\providecommand{\BIBentrySTDinterwordspacing}{\spaceskip=0pt\relax}
\providecommand{\BIBentryALTinterwordstretchfactor}{4}
\providecommand{\BIBentryALTinterwordspacing}{\spaceskip=\fontdimen2\font plus
\BIBentryALTinterwordstretchfactor\fontdimen3\font minus
  \fontdimen4\font\relax}
\providecommand{\BIBforeignlanguage}[2]{{%
\expandafter\ifx\csname l@#1\endcsname\relax
\typeout{** WARNING: IEEEtran.bst: No hyphenation pattern has been}%
\typeout{** loaded for the language `#1'. Using the pattern for}%
\typeout{** the default language instead.}%
\else
\language=\csname l@#1\endcsname
\fi
#2}}
\providecommand{\BIBdecl}{\relax}
\BIBdecl

\bibitem{Schmidhuber_2015}
J.~Schmidhuber, ``Deep learning in neural networks: An overview,'' \emph{Neural
  Networks}, vol.~61, pp. 85--117, 2015.

\bibitem{IsmailFawaz2019DeepLF}
H.~I. Fawaz, G.~Forestier, J.~Weber, L.~Idoumghar, and P.-A. Muller, ``Deep
  learning for time series classification: a review,'' \emph{Data Mining and
  Knowledge Discovery}, vol.~33, pp. 917--963, 2019.

\bibitem{Banko_2001}
M.~Banko and E.~Brill, ``Scaling to very very large corpora for natural
  language disambiguation,'' in \emph{AMACL}, 2001.

\bibitem{UCRArchive2018}
H.~A. Dau, E.~Keogh, K.~Kamgar, C.-C.~M. Yeh, Y.~Zhu, S.~Gharghabi, C.~A.
  Ratanamahatana, Yanping, B.~Hu, N.~Begum, A.~Bagnall, A.~Mueen, G.~Batista,
  and Hexagon-ML, ``The ucr time series classification archive,'' 2018,
  \url{https://www.cs.ucr.edu/~eamonn/time_series_data_2018/}.

\bibitem{iwana2021an}
B.~K. Iwana and S.~Uchida, ``An empirical survey of data augmentation for time
  series classification with neural networks,'' \emph{PLOS ONE}, 2021.

\bibitem{Shorten_2019}
C.~Shorten and T.~M. Khoshgoftaar, ``A survey on image data augmentation for
  deep learning,'' \emph{J. Big Data}, vol.~6, no.~1, 2019.

\bibitem{Niu_2019}
T.~Niu and M.~Bansal, ``Automatically learning data augmentation policies for
  dialogue tasks,'' in \emph{EMNLP-IJNLP}, 2019.

\bibitem{matsuo2021}
S.~Matsuo, S.~Uchida, and B.~K. Iwana, ``Self-augmented multi-modal feature
  embedding,'' in \emph{ICASSP}, 2021.

\bibitem{Cubuk_2019}
E.~D. Cubuk, B.~Zoph, D.~Mane, V.~Vasudevan, and Q.~V. Le, ``{AutoAugment}:
  Learning augmentation strategies from data,'' in \emph{CVPR}, 2019.

\bibitem{Jouppi2017IndatacenterPA}
N.~P. Jouppi, C.~Young, N.~Patil, and et.al, ``In-datacenter performance
  analysis of a tensor processing unit,'' \emph{2017 ACM/IEEE 44th Annual
  International Symposium on Computer Architecture (ISCA)}, pp. 1--12, 2017.

\bibitem{Tanner1987TheCO}
M.~A. Tanner and W.~H. Wong, ``The calculation of posterior distributions by
  data augmentation,'' \emph{Journal of the American Statistical Association},
  vol.~82, pp. 528--540, 1987.

\bibitem{Bishop1995TrainingWN}
C.~M. Bishop, ``Training with noise is equivalent to tikhonov regularization,''
  \emph{Neural Computation}, vol.~7, pp. 108--116, 1995.

\bibitem{an1996}
G.~An, ``The effects of adding noise during backpropagation training on a
  generalization performance,'' \emph{Neural Computation}, vol.~8, no.~3, pp.
  643--674, 1996.

\bibitem{Um_2017}
T.~T. Um, F.~M.~J. Pfister, D.~Pichler, S.~Endo, M.~Lang, S.~Hirche,
  U.~Fietzek, and D.~Kuli{\'{c}}, ``Data augmentation of wearable sensor data
  for parkinson's disease monitoring using convolutional neural networks,'' in
  \emph{ICMI}, 2017, pp. 216--220.

\bibitem{le2016data}
A.~{Le Guennec}, S.~Malinowski, and R.~Tavenard, ``Data augmentation for time
  series classification using convolutional neural networks,'' in
  \emph{IWAATD}, 2016.

\bibitem{rashid2019timewarp}
K.~M. Rashid and J.~Louis, ``Time-warping: A time series data augmentation of
  imu data for construction equipment activity identification,'' in
  \emph{Proceedings of the 36th International Symposium on Automation and
  Robotics in Construction (ISARC)}, M.~Al-Hussein, Ed.\hskip 1em plus 0.5em
  minus 0.4em\relax Banff, Canada: International Association for Automation and
  Robotics in Construction (IAARC), May 2019, pp. 651--657.

\bibitem{lim2019fast}
S.~Lim, I.~Kim, T.~Kim, C.~Kim, and S.~Kim, ``Fast autoaugment,''
  \emph{NeurIPS}, vol.~32, pp. 6665--6675, 2019.

\bibitem{Hataya_2020}
R.~Hataya, J.~Zdenek, K.~Yoshizoe, and H.~Nakayama, ``Faster {AutoAugment}:
  Learning augmentation strategies using backpropagation,'' in \emph{ECCV},
  2020.

\bibitem{Cubuk_2020}
E.~D. Cubuk, B.~Zoph, J.~Shlens, and Q.~V. Le, ``Randaugment: Practical
  automated data augmentation with a reduced search space,'' in \emph{CVPR
  Workshops}, 2020.

\bibitem{ho2019population}
D.~Ho, E.~Liang, X.~Chen, I.~Stoica, and P.~Abbeel, ``Population based
  augmentation: Efficient learning of augmentation policy schedules,'' in
  \emph{ICML}, 2019, pp. 2731--2741.

\bibitem{Li_2020}
R.~Li, X.~Li, P.-A. Heng, and C.-W. Fu, ``{PointAugment}: An auto-augmentation
  framework for point cloud classification,'' in \emph{CVPR}, 2020.

\bibitem{cheung2020modals}
T.-H. Cheung and D.-Y. Yeung, ``Modals: Modality-agnostic automated data
  augmentation in the latent space,'' in \emph{ICLR}, 2020.

\bibitem{shanmugam2020and}
D.~Shanmugam, D.~Blalock, G.~Balakrishnan, and J.~Guttag, ``When and why
  test-time augmentation works,'' \emph{arXiv preprint arXiv:2011.11156}, 2020.

\bibitem{Hu_2021}
T.-Y. Hu, A.~Shrivastava, J.-H.~R. Chang, H.~Koppula, S.~Braun, K.~Hwang,
  O.~Kalinli, and O.~Tuzel, ``{SapAugment}: Learning a sample adaptive policy
  for data augmentation,'' in \emph{ICASSP}, 2021.

\bibitem{fons2021adaptive}
E.~Fons, P.~Dawson, X.-j. Zeng, J.~Keane, and A.~Iosifidis, ``Adaptive
  weighting scheme for automatic time-series data augmentation,'' \emph{arXiv
  preprint arXiv:2102.08310}, 2021.

\bibitem{Wang_2017}
Z.~Wang, W.~Yan, and T.~Oates, ``Time series classification from scratch with
  deep neural networks: A strong baseline,'' in \emph{IJCNN}, 2017.

\bibitem{bai2018empirical}
S.~Bai, J.~Z. Kolter, and V.~Koltun, ``An empirical evaluation of generic
  convolutional and recurrent networks for sequence modeling,'' \emph{arXiv
  preprint arXiv:1803.01271}, 2018.

\bibitem{ioffe2015batch}
S.~Ioffe and C.~Szegedy, ``Batch normalization: Accelerating deep network
  training by reducing internal covariate shift,'' in \emph{ICML}, 2015, pp.
  448--456.

\bibitem{vandermaaten08a}
L.~van~der Maaten and G.~Hinton, ``Visualizing data using t-sne,'' \emph{J.
  Mach. Learn. Res.}, vol.~9, no.~86, pp. 2579--2605, 2008.

\end{thebibliography}

\end{document}